\documentclass[journal]{IEEEtran}
\ifCLASSINFOpdf
\else
\fi

\usepackage{epsfig, graphicx}
\usepackage{tikz}
\usepackage[colorlinks,
linkcolor=black,
anchorcolor=black,
citecolor=black
]{hyperref}
\usepackage{amsmath, amsthm, amssymb}
\usepackage{cite}
\usepackage{url}
\usepackage{diagbox}

\newtheorem{prop}{Proposition}
\newtheorem{theorem}{Theorem}
\newtheorem{lemma}{Lemma}
\newtheorem{definition}{Definition}
\newtheorem{corollary}{Corollary}
\usepackage{algorithm}
\usepackage{algorithmic}

\begin{document}
%
\title{Exploiting Low-Rank Tensor-Train Deep Neural Networks Based on Riemannian Gradient Descent With Illustrations of Speech Processing}
%
%
%

\author{Jun Qi,~\IEEEmembership{Member,~IEEE},
	Chao-Han Huck Yang,~\IEEEmembership{Student Member,~IEEE},
	Pin-Yu Chen,~\IEEEmembership{Member,~IEEE}, \\
 	Javier Tejedor	
\thanks{Jun Qi and Chao-Han Huck Yang are with the School of Electrical and Computer Engineering, Georgia Institute of Technology, Atlanta,
GA, 30332 USA e-mail: (qij41@gatech.edu, huckiyang@gatech.edu).}
\thanks{Pin-Yu Chen is with the IBM research, Yorktown Height, NY (e-mail: pin-yu.chen@ibm.com).}
\thanks{Javier Tejedor is with the Universidad San Pablo-CEU, CEU Universities, Madrid, Spain (javier.tejedornoguerales@ceu.es)}
}
\maketitle

\begin{abstract}
This work focuses on designing low complexity hybrid tensor networks by considering trade-offs between the model complexity and practical performance. Firstly, we exploit a low-rank tensor-train deep neural network (TT-DNN) to build an end-to-end deep learning pipeline, namely LR-TT-DNN. Secondly, a hybrid model combining LR-TT-DNN with a convolutional neural network (CNN), which is denoted as CNN+(LR-TT-DNN), is set up to boost the performance. Instead of randomly assigning large TT-ranks for TT-DNN, we leverage Riemannian gradient descent to determine a TT-DNN associated with small TT-ranks. Furthermore, CNN+(LR-TT-DNN) consists of convolutional layers at the bottom for feature extraction and several TT layers at the top to solve regression and classification problems. We separately assess the LR-TT-DNN and CNN+(LR-TT-DNN) models on speech enhancement and spoken command recognition tasks. Our empirical evidence demonstrates that the LR-TT-DNN and CNN+(LR-TT-DNN) models with fewer model parameters can outperform the TT-DNN and CNN+(TT-DNN) counterparts. 
\end{abstract}


%
\IEEEpeerreviewmaketitle

\section{Introduction}
\label{sec1}
%
%
%
%

\IEEEPARstart{T}{he} state-of-the-art deep learning systems for speech enhancement and spoken command recognition highly rely on the use of a great amount of training data and flexible access to centralized cloud services~\cite{yu2016automatic}. This results from the fact that an over-parameterized deep neural network (DNN) provides a simplified optimization landscape, which ensures local optimal points that are close to the global one~\cite{neyshabur2019towards}. Furthermore, a practical low-complexity speech enhancement system could create many new applications, such as the mobile-based audio de-noise and speech recognition platforms run on users' phones without sending requests to a remote server where a large deep learning model is set up. In doing so, a localized low-complexity speech processing system on local devices can maintain the baseline performance. Thus, it motivates us to design a low-complexity deep model without losing its capability in terms of representation and generalization powers~\cite{qi2019theory, qi2020analyzing, qi2022theoretical}. 

The approaches to shrinking deep learning models are divided into two categories: one refers to the use of new deep learning architectures, such as convolutional neural network (CNN) with various novel architectures~\cite{he2016deep}; another one denotes model pruning and sparseness methodologies~\cite{liu2018rethinking, zhu2017prune}, where either theoretically guaranteed pruning method or randomized lottery ticket hypothesis~\cite{malach2020proving, frankle2019stabilizing} can be used to attain a tiny machine learning system. In this work, we concentrate on the first aspect and investigate the deployment of low-rank tensor-train (TT) networks, where the experiments of speech enhancement and spoken command recognition (SCR) systems are illustrated to verify the effectiveness of our proposed models. 

TT denotes a tensor decomposition and can characterize the representation of a chain-like product of three-index core tensors for a high-order tensor~\cite{oseledets2011tensor}. In doing so, the memory overhead can be greatly reduced. Compared with other tensor decomposition methods~\cite{sidiropoulos2017tensor, kolda2009tensor} like Tucker decomposition (TD)~\cite{kim2007nonnegative} and CANDECOMP/PARAFAC decomposition~\cite{phan2011parafac, faber2003recent}, TT is a special case of a tree-structured tensor network and can be simply scaled to arbitrarily high-order tensors~\cite{novikov2015tensorizing}. Since TT provides a feed-forward DNN with compact multi-dimensional tensor formats, the TT representation for the DNN refers to a tensor-train deep neural network (TT-DNN), where each hidden layer of the DNN can be represented by the TT formats given randomly initialized TT-ranks. Our empirical study on TT-DNN for speech enhancement suggests that TT-DNN is capable of maintaining the DNN baseline performance, and it can significantly reduce the number of DNN parameters~\cite{qi2020tensor, Qi2020, qi2022exploring}. 

Although TT-DNN can construct a standalone speech enhancement~\cite{loizou2007speech} or SCR system~\cite{qi2021classical}, speech spectral features, like Mel-frequency cepstral coefficients (MFCC)~\cite{rabiner1993fundamentals} or Gammatone Features~\cite{qi2013auditory}, should be extracted and saved in additional hard disks beforehand. To realize an end-to-end training pipeline, our previous work~\cite{Qi2020} proposed a novel deep hybrid tensor-train model named CNN+(TT-DNN), which is composed of a CNN at the bottom and a TT-DNN at the top. The CNN model converts time-series signals into the corresponding spectral features, and the TT-DNN is used to further deal with regression or classification problems. The CNN+(TT-DNN) model takes advantage of the CNN for feature extraction and the TT-DNN for compact representation of the DNN such that an end-to-end deep hybrid tensor model can be built. 

Moreover, instead of randomly assigning TT-ranks to each hidden layer in the TT-DNN in our previous work, we can leverage Riemannian optimization to achieve a low-rank TT-DNN with fewer model parameters from a well-trained TT-DNN model. In more detail, given a set of large TT-ranks and a prescribed set of small TT-ranks, we employ the Riemannian gradient descent (RGD) algorithm to find a new TT-DNN model with the given small TT-ranks~\cite{bonnabel2013stochastic} from a well-trained TT-DNN with the large TT-ranks. The generated low-rank TT-DNN is named LR-TT-DNN in this work, which approximates the TT-DNN with the large TT-ranks. Besides, the LR-TT-DNN originates from a well-trained DNN, and it requires an additional fine-tuning process based on the algorithm of stochastic gradient descent (SGD)~\cite{bonnabel2013stochastic}. Furthermore, we deploy a hybrid model combining LR-TT-DNN with CNN, namely CNN+(LR-TT-DNN), to implement an end-to-end pipeline and boost the empirical performance of speech processing~\cite{Qi2020}. 

We separately employ speech enhancement and spoken command recognition experiments to assess the empirical performance of the LR-TT-DNN and CNN+(LR-TT-DNN) models, and we compare their performance with other TT models. The speech enhancement is a typical regression problem from a vector space to another vector space, and SCR represents a classification problem mapping from a vector space to a finite set of labels. 

The remainder of this paper is organized as follows: Section~\ref{sec2} summarizes the related work and our contribution to this work; Section~\ref{sec3} reviews the TT decomposition and tensor-train network (TTN); our proposed LR-TT-DNN model with the RGD algorithm is introduced in Section~\ref{sec4}, and the CNN+(LR-TT-DNN) model is presented in Section~\ref{sec5}. The related experiments of speech enhancement and SCR are shown in Section~\ref{sec6}. Finally, the paper is concluded in Section~\ref{sec7}. 

\section{Related Work and Our Contribution}
\label{sec2}
The technique of tensor-train decomposition (TTD) has been successfully applied in many machine learning tasks. In particular, Oseledets \emph{et al.}~\cite{oseledets2011tensor} first proposed the TTN model by employing TTD on neural networks with fully connected (FC) hidden layers. Then, it was extended to the CNN~\cite{garipov16ttconv} and recurrent neural network~\cite{yang2017tensor}. Yu \emph{et al.}~\cite{yu2017long} employed the TT model as a sequence-to-sequence dynamic model to forecast multivariate environmental data, and Izmailov \emph{et al.}~\cite{izmailov2018scalable} built scalable Gaussian processes with billions of inducing inputs by utilizing TTD. Another work makes use of TTN for channel estimation in wireless communications~\cite{zhang2021designing}. 

Moreover, Sidiropoulos \emph{et al.} investigated tensor factorization models and set up their relationship with signal processing and machine learning applications~\cite{sidiropoulos2017tensor}. Bacciu \emph{et al.} summarized tensor decompositions in deep learning, including TTD, Canonical Polyadic (CP) decomposition, and TD, and also compared their performance on neural model compression~\cite{bacciu2020tensor}. Furthermore, our previous empirical study~\cite{Qi2020} demonstrated that TD enables a compressed DNN model, but the model cannot guarantee the DNN performance. Since TD is a generalization of the CP decomposition (i.e., CP can be seen as TD with a super-diagonal core)~\cite{rabanser2017introduction}, this work only considers the comparison between TT and TD in terms of empirical performance. 

Although TTD on neural networks has exhibited excellent performance in many machine learning tasks, the TT-ranks are randomly assigned to the TT-DNN such that the model parameters of the TT-DNN can be randomly initialized followed by a thoroughly training via the SGD algorithm. On the other hand, a TT-DNN model with small prescribed TT-ranks could be iteratively found from the TT-DNN with a set of large TT-ranks. More specifically, this work concentrates on how to generate the LR-TT-DNN model based on the RGD algorithm. The RGD algorithm is related to the theory of SGD on Riemannian manifolds developed by~\cite{bonnabel2013stochastic}  and~\cite{luong2015effective}. Zhang \emph{et al.} proposed a fast stochastic optimization on Riemannian manifolds~\cite{zhang2016riemannian}. 

The contribution of this work can be summarized as follows: (1) leveraging the Riemannian optimization to create an LR-TT-DNN model with a set of prescribed small TT-ranks and illustrate its performance in speech enhancement and SCR tasks; (2) setting up hybrid deep tensor models and illustrates their performance in speech enhancement and SCR tasks. 

\section{Preliminaries}
\label{sec3}

\subsection{Notations}
We define $\mathbb{R}^{D}$ as a $D$-dimensional real coordinate space, and also assume $\mathbb{R}^{I_{1} \times I_{2} \times \cdot\cdot\cdot \times I_{K}}$ as the space of $K$-order tensors. $\nabla \mathcal{L}$ denotes the first-order gradient of the function $\mathcal{L}$, and $||\mathcal{X}||_{F}$ refers to the Frobenius norm of a tensor $\mathcal{X} \in \mathbb{R}^{I_{1} \times I_{2} \times \cdot\cdot\cdot \times I_{K}}$, which is given by Eq.~(\ref{eq:lll}). 
\begin{equation}
\label{eq:lll}
||\mathcal{X}||_{F} = \left[ \sum\limits_{i_{1}=1}^{I_{1}} \cdot\cdot\cdot \sum\limits_{i_{K}=1}^{I_{K}} \left(\mathcal{X}(i_{1}, ..., i_{K})\right)^{2} \right]^{\frac{1}{2}}. 
\end{equation}
Besides, the symbol $[K]$ means a set of integers $\{1, 2, ..., K\}$. 

\subsection{Tensor-Train Decomposition}

\begin{figure}
\centerline{\epsfig{figure=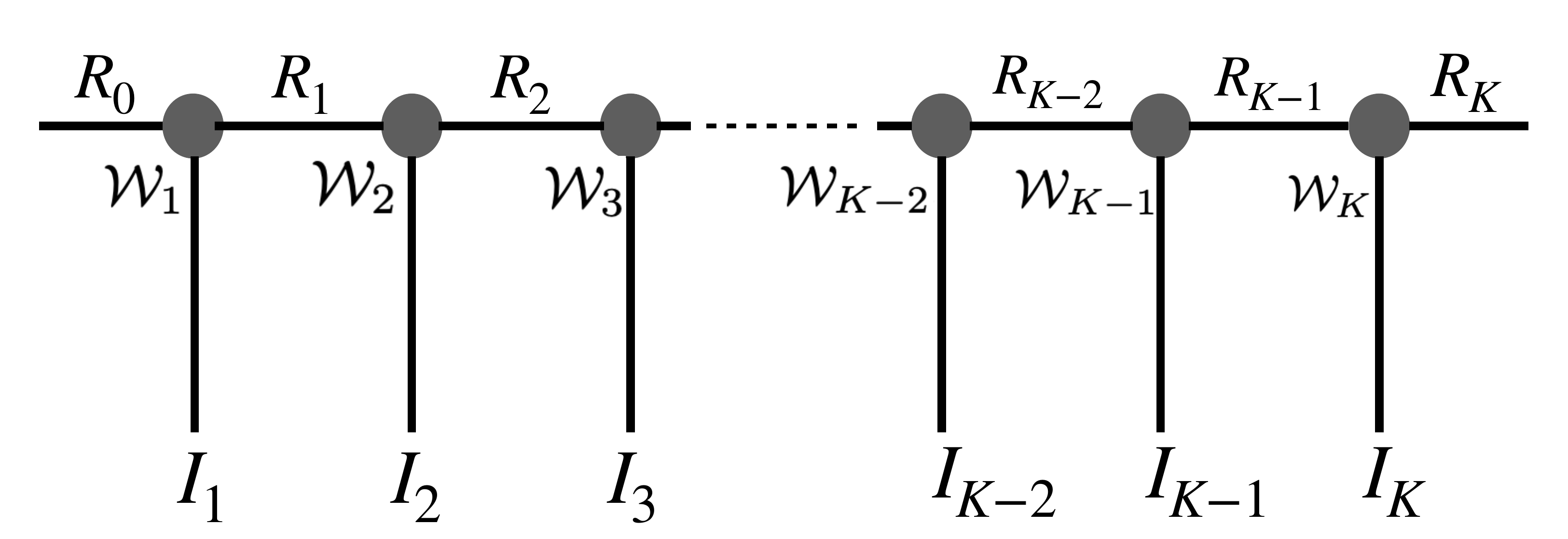, width=65mm}}
\caption{{\it An illustration of TTD, which is a tensor of order $K$ in TT format, where the core tensors are of order $3$. Given a set of TT-ranks $\{R_{0}, R_{1}, ..., R_{K}\}$, a circle represents a core tensor $\mathcal{W}_{k} \in \mathbb{R}^{I_{k} \times R_{k-1} \times R_{k}}$, and each line is associated with the dimension.}}
\label{fig:ttd}
\end{figure}

We first define some useful notations: $\mathbb{R}^{D}$ denotes a $D$-dimensional real space, $\mathbb{R}^{I_{1} \times I_{2} \times \cdot\cdot\cdot I_{K}}$ is a space of $K$-order tensors, $\mathcal{W} \in \mathbb{R}^{I_{1} \times \cdot\cdot\cdot \times I_{K}}$ is a $K$-dimensional tensor, and $\textbf{W} \in \mathbb{R}^{I_{1} \times I_{2}}$ is an $I_{1} \times I_{2}$ matrix.

As shown in Figure~\ref{fig:ttd}, the technique of TTD assumes that given a set of TT-ranks $\{R_{0}, R_{1}, ..., R_{K}\}$, a $K$-order tensor $\mathcal{W}\in \mathbb{R}^{I_{1}\times I_{2}\times \cdot\cdot\cdot\times I_{K}}$ is factorized into the multiplication of 3-order tensors $\mathcal{W}_{k} \in \mathbb{R}^{I_{k} \times R_{k-1} \times R_{k}}$. More specifically, given a set of indices $\{i_{1}, i_{2}, ..., i_{K}\}$, $\mathcal{W}(i_{1}, i_{2}, ..., i_{K})$ is decomposed as Eq.~(\ref{eq:tt}). 
\begin{equation}
\label{eq:tt}
\mathcal{W}(i_{1}, i_{2}, ..., i_{K}) = \prod\limits_{k=1}^{K} \mathcal{W}_{k}(i_{k}),
\end{equation}
where $\forall i_{k} \in I_{k}$, $\mathcal{W}_{k} \in \mathbb{R}^{I_{k} \times R_{k-1}\times R_{k}}$ and $\mathcal{W}_{k}(i_{k}) \in \mathbb{R}^{R_{k-1} \times R_{k}}$. Since $R_{0} = R_{K} = 1$, the term $\prod_{k=1}^{K} \mathcal{W}_{k}(i_{k})$ is a scalar value.

One example of TTD is shown as follows: for a $3$-order tensor $\mathcal{W}(i_{1}, i_{2}, i_{3}) = i_{1} + i_{2} + i_{3}$, given the set of TT-ranks $\{1, 2, 2, 1\}$, the use of TTD on $\mathcal{W}$ outputs $3$ core tensors as Eq.~(\ref{eq:2}). 
\begin{equation}
\label{eq:2}
\mathcal{W}_{1}[i_{1}] := [ \begin{matrix} i_{1} & 1  \end{matrix}],  \mathcal{W}_{2}[i_{2}] := \left[\begin{matrix} 1 & 0 \\ i_{2} & 1 \end{matrix}\right], \mathcal{W}_{3}[i_{3}] := \left[ \begin{matrix} 1 \\ i_{3} \end{matrix} \right], 
\end{equation}
which is derived from Eq.~(\ref{eq:ww}). 
\begin{equation}
\label{eq:ww}
\mathcal{W}(i_{1}, i_{2}, i_{3}) = [\begin{matrix} i_{1} & 1  \end{matrix}] \left[\begin{matrix} 1 & 0 \\ i_{2} & 1 \end{matrix}\right] \left[ \begin{matrix} 1 \\ i_{3} \end{matrix} \right] = i_{1} + i_{2} + i_{3}.
\end{equation}

\subsection{Tensor-Train Deep Neural Network}
\begin{figure}
\centerline{\epsfig{figure=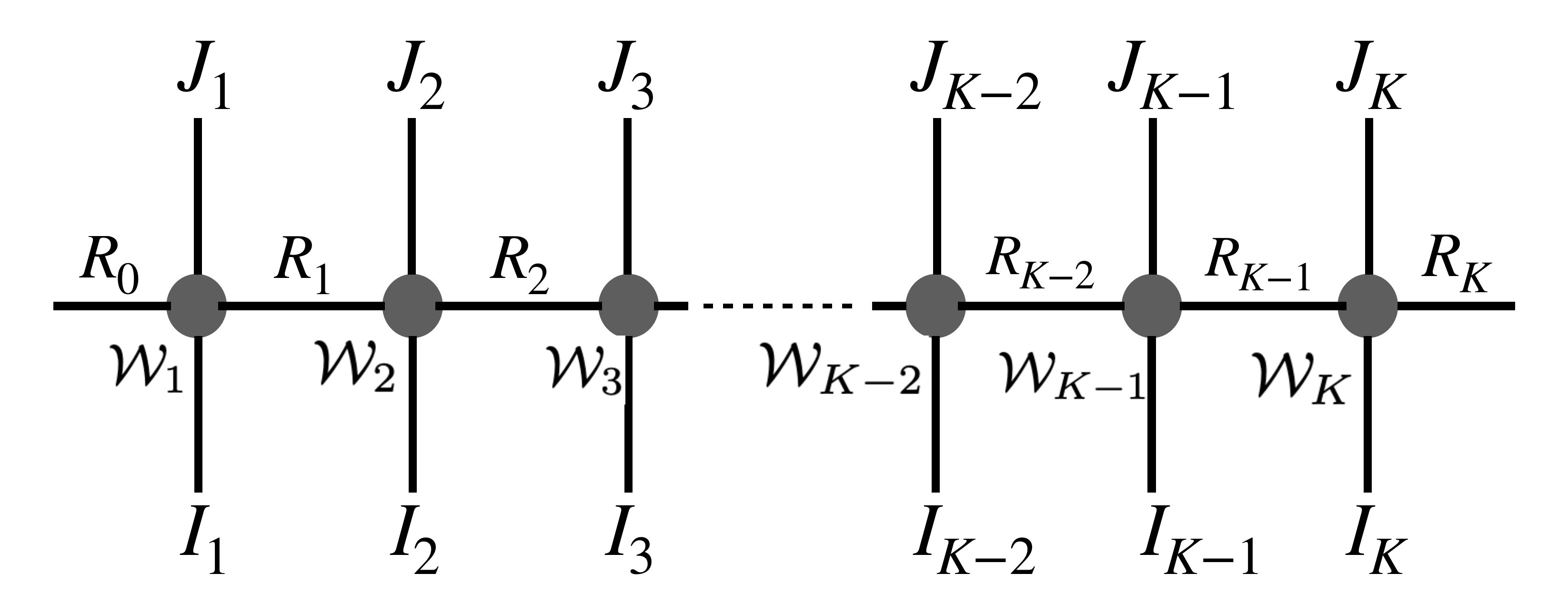, width=65mm}}
\caption{{\it An illustration of TTN, which is a tensor of order $K$ in TT format, where the core tensors are of order $4$. Given a set of TT-ranks $\{R_{0}, R_{1}, ..., R_{K}\}$, a circle represents a core tensor $\mathcal{W}_{k} \in \mathbb{R}^{I_{k} \times J_{k} \times R_{k-1} \times R_{k}}$, and each line is associated with the dimension.}}
\label{fig:ttnn}
\end{figure}

A TTN denotes a TT representation of a feed-forward neural network with a FC hidden layer. In more detail, for an input tensor $\mathcal{X} \in \mathbb{R}^{I_{1} \times I_{2} \times \cdot\cdot\cdot \times I_{K}}$ and an output tensor $\mathcal{Y} \in \mathbb{R}^{J_{1} \times J_{2} \times \cdot\cdot\cdot \times J_{K}}$, we get Eq.~(\ref{eq:yy}) as follows:
\begin{equation}
\label{eq:yy}
\begin{split}
&\hspace{4mm} \mathcal{Y}(j_{1}, j_{2}, ..., j_{K})  \\
&= \sum\limits_{i_{1}=1}^{I_{1}} \cdot\cdot\cdot \sum\limits_{i_{K} = 1}^{I_{K}} \mathcal{W}((i_{1}, j_{1}), ..., (i_{K}, j_{K})) \cdot \mathcal{X}(i_{1}, i_{2}, ..., i_{K}) \\
&= \sum\limits_{i_{1}=1}^{I_{1}} \cdot\cdot\cdot \sum\limits_{i_{K} = 1}^{I_{K}} \left( \prod\limits_{k=1}^{K} \mathcal{W}_{k}(i_{k}, j_{k}) \right) \cdot \prod\limits_{k=1}^{K} \mathcal{X}_{k}(i_{k})		 \\
&= \prod\limits_{k=1}^{K} \left( \sum\limits_{i_{k}=1}^{I_{K}} \mathcal{W}_{k}(i_{k}, j_{k}) \cdot \mathcal{X}_{k}(i_{k})  \right)		\\
&= \prod\limits_{k=1}^{K} \mathcal{Y}_{k}(j_{k}),
\end{split}
\end{equation}
where $\mathcal{X}_{k}(i_{k}) \in \mathbb{R}^{R_{k-1} \times R_{k}}$, and $\mathcal{Y}_{k}(j_{k}) \in \mathbb{R}^{R_{k-1} \times R_{k}}$ which results in a scalar $\prod_{k=1}^{K} \mathcal{Y}_{k}(j_{k})$ because of the ranks $R_{0} = R_{K} = 1$; $\mathcal{W}((i_{1}, j_{1}), (i_{2}, j_{2}), ..., (i_{K}, j_{K}))$ is closely associated with $\mathcal{W}(m_{1}, m_{2}, ..., m_{K})$ as defined in Eq.~(\ref{eq:tt}), if the indexes $m_{k} = i_{k} \times j_{k}, k\in [K]$ are set. As shown in Figure~\ref{fig:ttnn}, given the TT-ranks $\{R_{0}, R_{1}, ..., R_{K}\}$, the multi-dimensional tensor $\mathcal{W}$ is decomposed into the multiplication of $4$-order tensors $\mathcal{W}_{k} \in \mathbb{R}^{I_{k} \times J_{k} \times R_{k-1} \times R_{k}}$. Besides, to configure a TTN, the ReLU activation is applied to $\mathcal{Y}(j_{1}, j_{2}, ..., j_{K})$ as Eq.~(\ref{eq:6}).
\begin{equation}
\label{eq:6}
\begin{split}
\mathcal{\hat{Y}}(j_{1}, j_{2}, ..., j_{K}) &= \text{ReLU}(\mathcal{Y}(j_{1}, j_{2}, ..., j_{K})) \\
&= \text{ReLU}\left(\prod\limits_{k=1}^{K} \mathcal{Y}_{k}(j_{k})\right).
\end{split}
\end{equation}

\begin{figure}
\centerline{\epsfig{figure=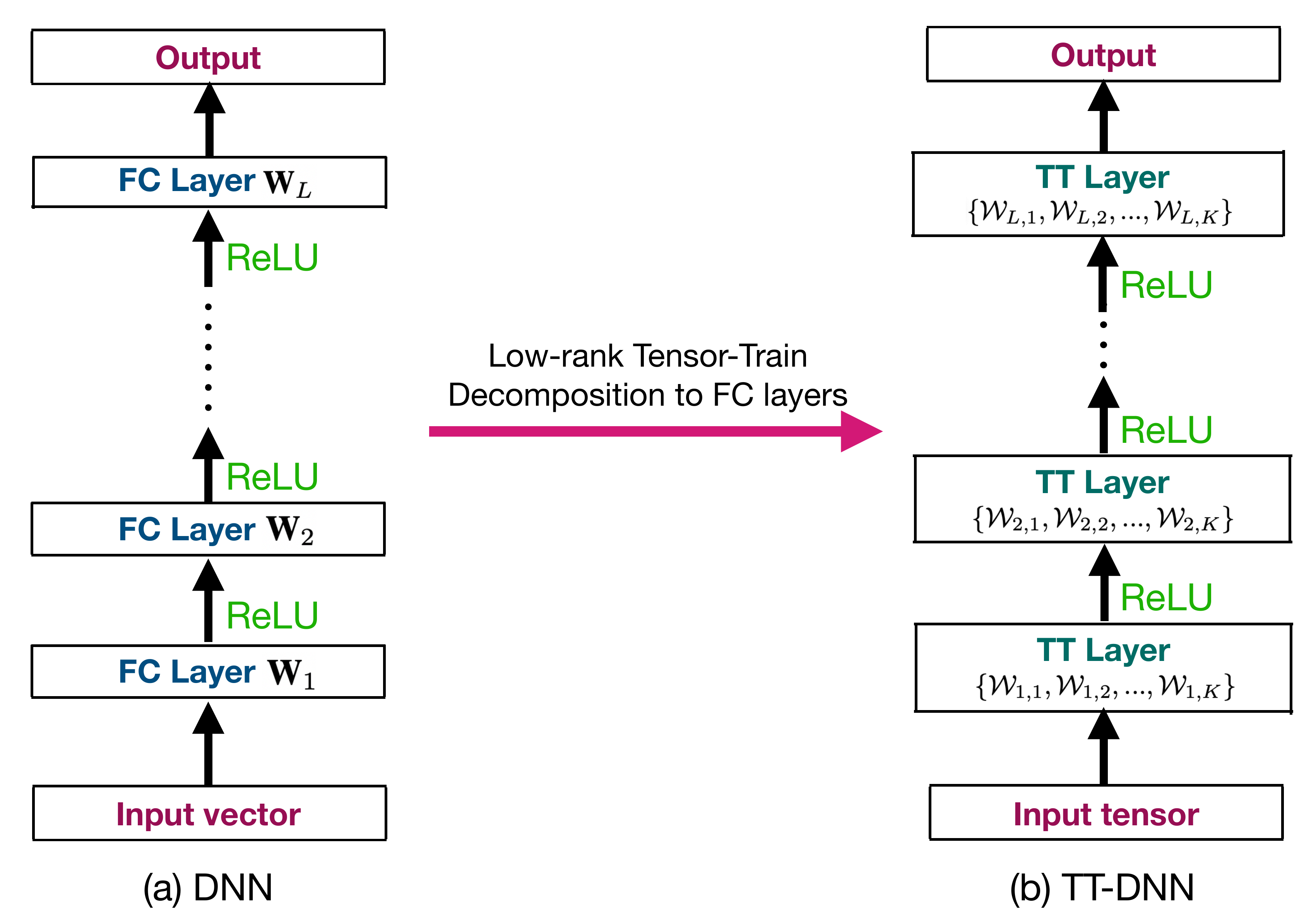, width=80mm}}
\caption{{\it Converting DNN into TT-DNN. Each DNN weight $\textbf{W}_{l}$ is connected to $K$ core tensors $\mathcal{W}_{l, 1}, \mathcal{W}_{l, 2}, ..., \mathcal{W}_{l, K}$. }}
\label{fig:ttn}
\end{figure}

Then, TTN can be generalized to a deeper architecture and is closely associated with the TT representation for DNN, namely TT-DNN. Although the TT-DNN model can be set up and trained from scratch by using the SGD algorithm, Figure~\ref{fig:ttn} illustrates that a DNN model can be transformed into a TT-DNN structure, where all FC hidden layers can be represented as the corresponding TT layers. More specifically, $\forall k\in [K]$ and $\forall l\in [L]$, the DNN matrix $\textbf{W}_{l} \in \mathbb{R}^{J_{l} \times I_{l}}$ can be decomposed into $K$ core tensors $\{\mathcal{W}_{l, 1}, \mathcal{W}_{l, 2}, ..., \mathcal{W}_{l, K}\}$, where $\mathcal{W}_{l, k} \in \mathbb{R}^{J_{l,k} \times I_{l,k} \times R_{k-1} \times R_{k}}$, $J_{l}=J_{l, 1} \times J_{l, 2} \times \cdot\cdot\cdot \times J_{l, K}$ and $I_{l} = I_{l, 1} \times I_{l, 2} \times \cdot\cdot\cdot \times I_{l, K}$.

The TTD admits a TT-DNN with much fewer number of model parameters than the related DNN. More specifically, a DNN with $\sum_{l=1}^{L}J_{l} I_{l}$ parameters can be converted to a TT-DNN with fewer model parameters such as $\sum_{l=1}^{L}\sum_{k=1}^{K} J_{l, k}I_{l, k} R_{k-1}R_{k}$.

\section{Low-Rank Tensor-Train Deep Neural Network}
\label{sec4}
In Section~\ref{sec3}, given a set of TT-ranks, we can initialize a TT-DNN model associated with a fixed tensor manifold corresponding to the given TT-ranks, where the model parameters of the TT-DNN are trained by applying the SGD algorithm from randomized initialized weights. On the other hand, the TT-DNN model can also be built according to Figure~\ref{fig:ttn}. From the TT-DNN model in Figure~\ref{fig:ttn}, an LR-TT-DNN model can be attained by approximating that TT-DNN model with a set of large TT-ranks by applying the RGD algorithm which iteratively finds the TT-DNN with the given small TT-ranks. In this section, we first introduce the RGD algorithm, and then we theoretically analyze its convergence and performance. 

\subsection{Riemannian Gradient Descent}

The objective of Riemannian optimization aims at designing a framework to solve an optimization problem with the constraint that the solution belongs to manifolds with a given set of small TT-ranks. The related Riemannian optimization is formulated as follows: given a TT tensor $\mathcal{W} = \{\mathcal{W}_{1}, \mathcal{W}_{2}, ..., \mathcal{W}_{K}\}$ with a set of prescribed large TT-ranks $R = \{R_{1}, R_{2}, ..., R_{K}\}$, we aim at finding the low-rank TT tensor $\hat{\mathcal{W}} = \{\hat{\mathcal{W}}_{1}, \hat{\mathcal{W}}_{2}, ..., \hat{\mathcal{W}}_{K}\}$ with a set of given small TT-ranks $\hat{R} = \{\hat{R}_{1}, \hat{R}_{2}, ..., \hat{R}_{K}\}$ being close to $\mathcal{W}$, which is mathematically described as Eq.~(\ref{eq:loww}). 
\begin{equation}
\label{eq:loww}
\begin{split}
& \hspace{15mm} \min\limits_{\mathcal{\hat{W}}} \mathcal{L}(\mathcal{W}, \mathcal{\hat{W}}) \\
&s.t.,  \text{TT-rank}(\mathcal{\hat{W}}) = \hat{R},  \hspace{1mm} \text{TT-rank}(\mathcal{W}) = R,
\end{split}
\end{equation}
where $\text{TT-rank}(\mathcal{W}) = R$ means that $\forall k\in [K]$, the $k$-th TT-rank of $\mathcal{W}$ is equal to $R_{k}$. 

Besides, the objective function $\mathcal{L}(\mathcal{W}, \mathcal{\hat{W}})$ is decomposed as Eq.~(\ref{eq:highh}). 
\begin{equation}
\label{eq:highh}
\mathcal{L}(\mathcal{W}, \mathcal{\hat{W}}) = \sum\limits_{k=1}^{K} \mathcal{L}(\mathcal{W}_{k}, \mathcal{\hat{W}}_{k}). 
\end{equation}

\begin{figure}[htp]
\centerline{\epsfig{figure=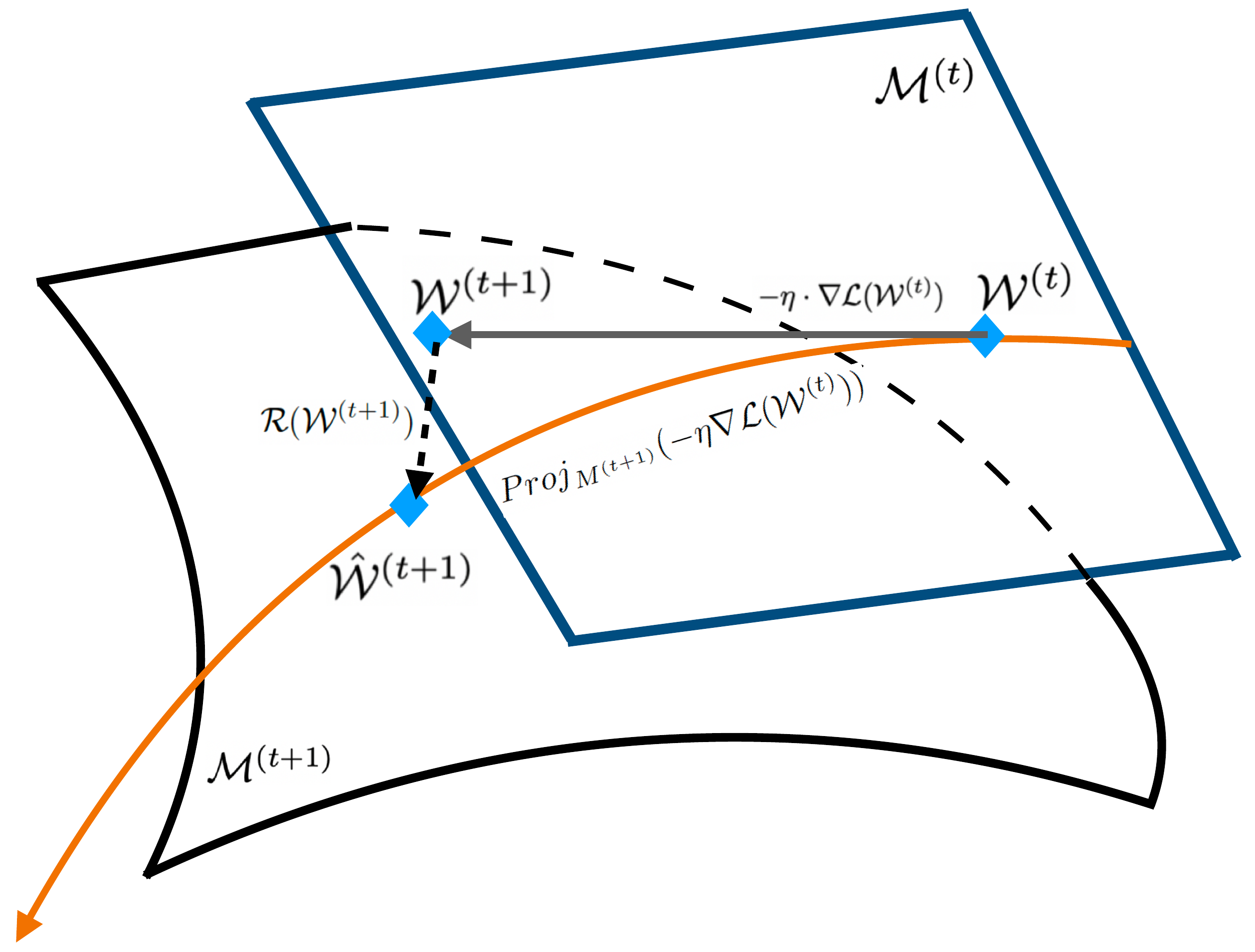, width=85mm}}
\caption{{\it An illustration of Riemannian gradient descent.}}
\label{fig:manifold}
\end{figure}

Moreover, the core tensors with element-wise fixed TT-ranks form a manifold. The optimization problem as shown in Eq.~(\ref{eq:loww}) can be solved by utilizing the RGD algorithm, which iteratively finds the low-rank manifold of $\mathcal{\hat{W}}$ from the original $\mathcal{W}$. Given a function $\mathcal{L}$ (e.g., $||\cdot||_{F}$) on a manifold $\mathcal{M}^{(t)}$, the updated TT tensor $\mathcal{W}^{(t+1)}$ relies on $\mathcal{W}^{(t)}$ as Eq.~(\ref{eq:www}). 
\begin{equation}
\label{eq:www}
\begin{split}
&\mathcal{W}^{(t+1)} = \mathcal{W}^{(t)} -\eta \nabla \mathcal{L}(\mathcal{W}^{(t)})	  \\
&\hspace{10mm} \mathcal{\hat{W}}^{(t+1)} = \mathcal{R}(\mathcal{W}^{(t+1)}),
\end{split}
\end{equation}
where $\eta$ is a learning rate, $\mathcal{R}$ is a retraction, and the operator ${Proj}_{M^{(t+1)}}(-\eta \nabla \mathcal{L}(\mathcal{W}^{(t)}))$ in Figure~\ref{fig:manifold} is a projection from the gradient descent $-\eta \nabla \mathcal{L}(\mathcal{W}^{(t)})$ onto a new tangent space $\mathcal{M}^{(t+1)}$ where $\mathcal{\hat{W}}^{(t+1)}$ is a low-rank TT representation of $\mathcal{M}^{(t)}$. The RGD process is illustrated in Figure~\ref{fig:manifold}: a TT tensor $\mathcal{W}^{(t)}$ starts at the manifold $\mathcal{M}^{(t)}$ and moves to $\mathcal{W}^{(t+1)}$ along the gradient descent $-\eta \nabla \mathcal{L}(\mathcal{W}^{(t)})$. The updated TT tensor $\mathcal{\hat{W}}^{(t+1)}$ with lower TT-ranks, which corresponds to the manifold $\mathcal{M}^{(t+1)}$, can be attained based on the retraction algorithm as shown in Algorithm~\ref{alg:rounding}. The retraction algorithm iteratively attempts to find a new manifold with lower TT-ranks. The main procedure of the retraction algorithm refers to a singular value decomposition (SVD) and the low-rank matrices $\hat{U}_{k}$, $\hat{\Sigma}_{k}$, and $\hat{V}_{k}$ are formed based on the updated TT rank $\hat{r}_{k+1}$. The low-rank matrix $\hat{A}_{k}$ is reshaped into a tensor format as $\mathcal{\hat{W}}_{k}$. 

\begin{algorithm}[tb]
   \caption{The Retraction Algorithm}
   \label{alg:rounding}
\begin{algorithmic}
   \STATE {\bfseries 1.} Give the core tensors $\{ \mathcal{W}_{1}, \mathcal{W}_{2}, ..., \mathcal{W}_{K} \}$ with TT-ranks $r = \{r_{1}, r_{2}, ..., r_{K}\}$, and a prescribed maximum rank $r_{\max}$. 
   \STATE {\bfseries 2.} Initialize a new set of TT-ranks $\hat{r} = \{1, 1, ..., 1\}$. 
   \STATE {\bfseries 3.} For $k = 1:K$, 
   \STATE {\bfseries 4.} \hspace{4mm} $\textbf{A}_{k} = \text{reshape}\hspace{1mm} \mathcal{W}_{k} \in \mathbb{R}^{r_{k} \times n_{k} \times m_{k} \times r_{k+1}}$ by the \\ \hspace{17mm} shape $(r_{k}\cdot n_{k} \cdot m_{k}) \times r_{k+1}$. 
   \STATE {\bfseries 5.} For $k = 1:K$, 
   \STATE {\bfseries 6.} \hspace{4mm} $\hat{r}_{k+1} = \min(r_{\max}, n_{k}\cdot m_{k} \cdot \hat{r}_{k+1}, \frac{r_{k} \cdot r_{k+1}}{\hat{r}_{k+1}})$. 
   \STATE {\bfseries 7.} \hspace{4mm} $(U_{k}, \Sigma_{k}, V_{k}) = \text{SVD}(\textbf{A}_{k})$. 
   \STATE {\bfseries 8.} \hspace{4mm} $\hat{U}_{k} = U_{k}[:, 1:\hat{r}_{k+1}]$, $\hat{\Sigma}_{k} = \Sigma_{k}[1:\hat{r}_{k+1}, 1:\hat{r}_{k+1}]$, \\
   \hspace{8mm} $\hat{V}_{k} = V_{k}[1:\hat{r}_{k+1}, :]$. 
   \STATE {\bfseries 9.} \hspace{4mm} $\hat{\textbf{A}}_{k} = \hat{U}_{k} \hat{\Sigma}_{k} \hat{V}_{k}$. 
   \STATE {\bfseries 10.} \hspace{2mm} $\mathcal{\hat{W}}_{k} \leftarrow$ reshape $\hat{\textbf{A}}_{k}$ as $\hat{r}_{k} \times n_{k} \times m_{k} \times \hat{r}_{k+1}$. 
   \STATE {\bfseries 11.} Return the updated core tensors $\{\mathcal{\hat{W}}_{1}, \mathcal{\hat{W}}_{2}, ..., \mathcal{\hat{W}}_{K}\}$ \\
   \hspace{5mm} and the updated lower ranks $\hat{r} = \{\hat{r}_{0}, \hat{r}_{1}, ..., \hat{r}_{K}\}$. 
\end{algorithmic}
\end{algorithm}

\subsection{Convergence Analysis}

\begin{theorem}
\label{thm:thm1}
Given an $H$-smooth and convex loss function $\mathcal{L}$, the algorithm of Riemannian gradient descent as Eq.~(\ref{eq:www}) ensures both Eq. (\ref{eq:ineq1}) and Eq. (\ref{eq:ineq2}). 
\begin{equation}
\label{eq:ineq1}
\max\limits_{t\in [T]}  || \nabla \mathcal{L}(\mathcal{W}^{(t)}) ||^{2} \le \frac{2}{\eta(2 - H\eta)} \left( \mathcal{L}(\mathcal{W}^{(0)}) - \mathcal{L}(\mathcal{W}^{*}) \right), 
\end{equation}
\begin{equation}
\label{eq:ineq2}
\min\limits_{t\in [T]}  || \nabla \mathcal{L}(\mathcal{W}^{(t)}) ||^{2}  \le \frac{2}{\eta(2 - H\eta)T} \left( \mathcal{L}(\mathcal{W}^{(0)}) - \mathcal{L}(\mathcal{W}^{*}) \right), 
\end{equation}
where $\mathcal{W}^{*}$ and $\mathcal{W}^{(0)}$ separately denote the optimal and initial TT tensors, $\eta \in (0, \frac{2}{H})$ denotes the learning rate, and $T$ refers to the number of steps. 
\end{theorem}

\begin{proof}
First, we define 
\begin{equation*}
\begin{split}
\mathcal{T}_{\mathcal{W}^{(t)}}(\mathcal{W}^{(t+1)}) &= \mathcal{W}^{(t+1)} - \mathcal{W}^{(t)} = -\eta \cdot \nabla \mathcal{L}(\mathcal{W}^{(t)}). 
\end{split}
\end{equation*}

Since $\mathcal{L}$ is an $H$-smooth function, Eq.~(\ref{eq:ineq3}) stands. 
\begin{equation}
\label{eq:ineq3}
\begin{split}
&\hspace{5mm} \mathcal{L}(\mathcal{W}^{(t+1)}) \\
&\le \mathcal{L}(\mathcal{W}^{(t)}) + \left\langle \nabla \mathcal{L}(\mathcal{W}^{(t)}), \mathcal{T}_{\mathcal{W}^{(t)}}(\mathcal{W}^{(t+1)}) \right\rangle \\
&\hspace{40mm} + \frac{H}{2} || \mathcal{T}_{\mathcal{W}^{(t)}}(\mathcal{W}^{(t+1)}) ||^{2} 	 \\
&\le \mathcal{L}(\mathcal{W}^{(t)}) - \eta  || \nabla \mathcal{L}(\mathcal{W}^{(t)}) ||^{2}  + \frac{H \eta^{2}}{2}  || \nabla \mathcal{L}(\mathcal{W}^{(t)}) ||^{2}.   \\
\end{split}
\end{equation}

After rearrangement in Eq.~(\ref{eq:ineq3}), we obtain Eq.~(\ref{eq:ineq4}) as:
\begin{equation}
\label{eq:ineq4}
||\nabla \mathcal{L}(\mathcal{W}^{(t)}) ||^{2} \le \frac{2}{\eta (2 - H \eta)} \left(	\mathcal{L}(\mathcal{W}^{(t)}) - \mathcal{L}(\mathcal{W}^{(t+1)})	\right).
\end{equation}

Summing up Eq.~(\ref{eq:ineq4}) from $t=0$ to $T-1$, we obtain Eq.~(\ref{eq:ineq5}) as:
\begin{equation}
\label{eq:ineq5}
\begin{split}
\sum\limits_{t=0}^{T-1}  || \nabla \mathcal{L}(\mathcal{W}^{(t)}) ||^{2}   &\le  \frac{2}{\eta(2 - H\eta)} \left( \mathcal{L}(\mathcal{W}^{(0)}) - \mathcal{L}(\mathcal{W}^{(T)}) \right)	\\
&\le \frac{2}{\eta(2 - H\eta)} \left( \mathcal{L}(\mathcal{W}^{(0)}) - \mathcal{L}(\mathcal{W}^{*}) \right), 	
 \end{split}
\end{equation}
where $\mathcal{W}^{*}$ denotes the optimal tensor such that $\mathcal{L}(\mathcal{W}^{*}) \le \mathcal{L}(\mathcal{W}^{(t)}), \forall t \in [0, T-1]$, and $0 < \eta < \frac{2}{H}$ ensures a positive upper bound. 

Finally, we separately derive Eq.~(\ref{eq:b1}) and Eq.~(\ref{eq:b2}) as follows:
\begin{equation}
\label{eq:b1}
\max\limits_{t\in[T]} || \nabla \mathcal{L}(\mathcal{W}^{(t)}) ||^{2} \le \sum\limits_{t=0}^{T-1}  || \nabla \mathcal{L}(\mathcal{W}^{(t)}) ||^{2}, 
\end{equation}
\begin{equation}
\label{eq:b2}
T \min\limits_{t\in[T]} || \nabla \mathcal{L}(\mathcal{W}^{(t)}) ||^{2} \le \sum\limits_{t=0}^{T-1}  || \nabla \mathcal{L}(\mathcal{W}^{(t)}) ||^{2}, 
\end{equation}
so that we attain Eq.~(\ref{eq:ineq1}) and Eq.~(\ref{eq:ineq2}).

\end{proof}

\subsection{Performance Analysis}

\begin{lemma}
\label{lem1}
The loss function $\mathcal{L}$ is $\tau$-gradient dominated if $\mathcal{W}^{*}$ is a global minimizer of $\mathcal{L}$ and for every $\mathcal{W}$, Eq.~(\ref{eq:performance}) stands. 
\begin{equation}
\label{eq:performance}
\mathcal{L}(\mathcal{W}) - \mathcal{L}(\mathcal{W}^{*}) \le \tau ||\nabla \mathcal{L}(\mathcal{W}) ||^{2}. 
\end{equation}
\end{lemma}

\begin{prop}
\label{prop1}
For an $H$-smooth and convex loss function $\mathcal{L}$ with the $\tau$-gradient dominated property, the algorithm of Riemannian gradient descent guarantees the performance at the convergence tensor $\mathcal{W}^{(\alpha)}$ as:
\begin{equation}
\label{eq:ens}
\begin{split}
\mathcal{L}(\mathcal{W}^{(\alpha)}) - \mathcal{L}(\mathcal{W}^{*}) &\le \tau || \nabla \mathcal{L}(\mathcal{W}^{(\alpha)}) ||^{2} \\
&\le \frac{2}{\eta(2 - H\eta)T} \left( \mathcal{L}(\mathcal{W}^{(0)}) - \mathcal{L}(\mathcal{W}^{*}) \right),
\end{split}
\end{equation}
where $\mathcal{W}^{(\alpha)} \leftarrow \arg\min\limits_{t\in [T]}   || \nabla \mathcal{L}(\mathcal{W}^{(t)}) ||^{2} $. 
\end{prop}

\section{CNN + Low-Rank Tensor-Train Deep Neural Network}
\label{sec5}
\begin{figure}[htp]
\centerline{\epsfig{figure=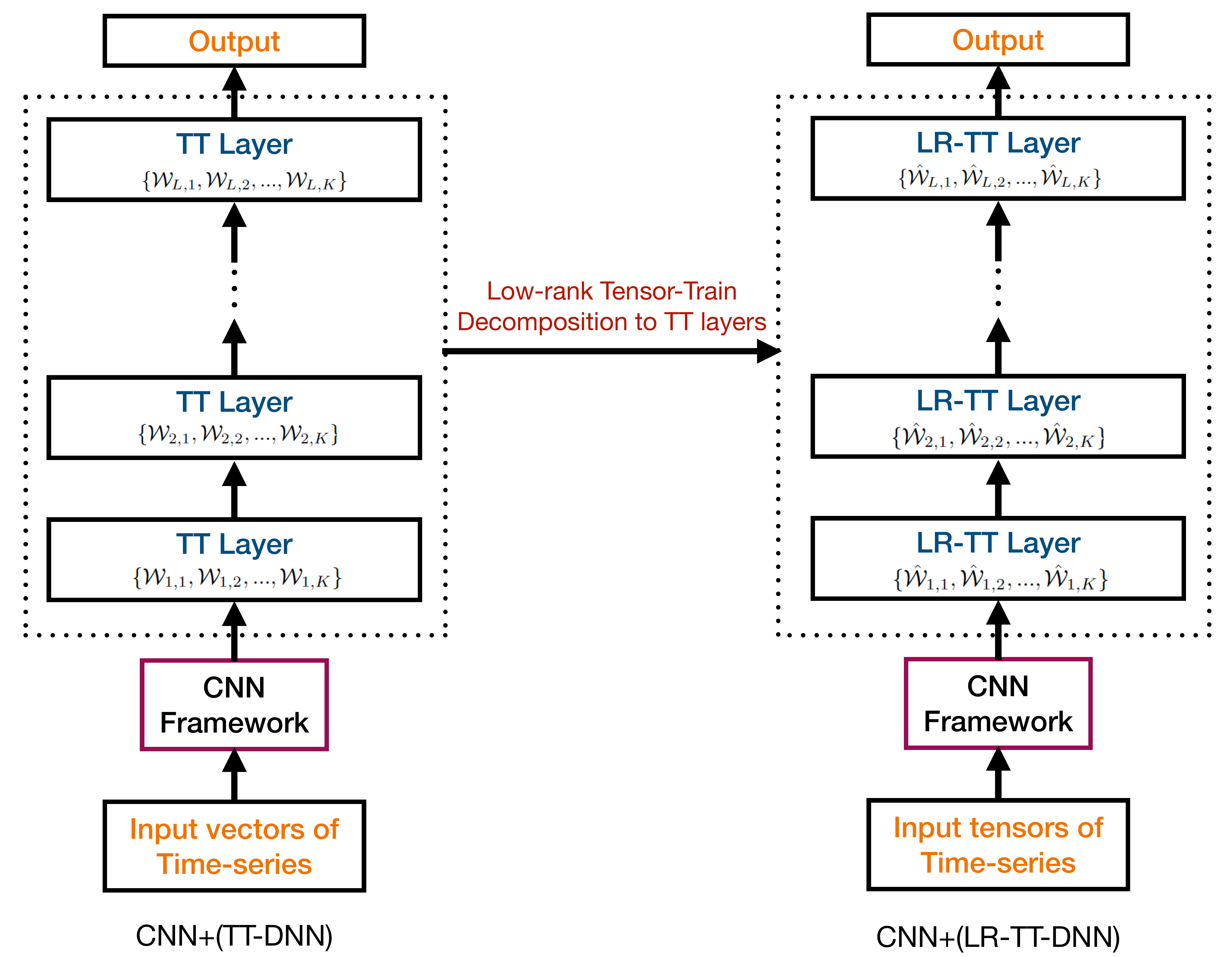, width=75mm}}
\caption{{\it An illustration of converting a CNN+(TT-DNN) model into a CNN+(LR-TT-DNN) model.}}
\label{fig:rie}
\end{figure}

Figure~\ref{fig:rie} shows that a CNN+(LR-TT-DNN) model can be attained from a well-trained CNN+(TT-DNN) model, where each TT layer is changed to an LR-TT layer which results in an LR-TT-DNN on top of the CNN. The attained LR-TT-DNN model can be further fine-tuned by using the SGD algorithm. 

As for the setup of the LR-TT-DNN model in our experiments, the Frobenius norm $||\cdot||_{F}$ is employed, and the learning rate $\eta$ is set to $1.0$. The ReLU activation is imposed upon each LR-TT layer. Since $||\cdot||_{F}$ is a smooth and convex function, we show that $||\cdot||_{F}$ is $\tau$-gradient dominated (see Appendix~\ref{app1}). 
\begin{corollary}
\label{cor1}
Frobenius norm $||\cdot||_{F}$ is a $1$-gradient dominated.
\end{corollary}

\begin{proof}
Suppose $\mathcal{L}(\mathcal{W}) = ||\mathcal{W}||_{F}$, and assume $\mathcal{W}^{*}$ as the optimal tensor, so that Eq.~(\ref{eq:proof1}) stands. 
\begin{equation}
\label{eq:proof1}
\begin{split}
&\hspace{4mm} \mathcal{L}(\mathcal{W}) - \mathcal{L}(\mathcal{W}^{*})	 \\
&= ||\mathcal{W}||_{F} - ||\mathcal{W}^{*}||_{F} 	\\
&= \frac{\sum\limits_{i_{1}, ..., i_{K}} (\mathcal{W}(i_{1}, ..., i_{K}))^{2} - \sum\limits_{i_{1}, ..., i_{K}} (\mathcal{W}^{*}(i_{1}, ..., i_{K}))^{2} }{\sqrt{\sum\limits_{i_{1}, ..., i_{K}} (\mathcal{W}(i_{1}, ..., i_{K}))^{2}} + \sqrt{\sum\limits_{i_{1}, ..., i_{K}} (\mathcal{W}^{*}(i_{1}, ..., i_{K}))^{2}}} 	\\
&\le \frac{\sum\limits_{i_{1}, ..., i_{K}} (\mathcal{W}(i_{1}, ..., i_{K}))^{2} }{\sqrt{\sum\limits_{i_{1}, ..., i_{K}} (\mathcal{W}(i_{1}, ..., i_{K}))^{2}} + \sqrt{\sum\limits_{i_{1}, ..., i_{K}} (\mathcal{W}^{*}(i_{1}, ..., i_{K}))^{2}}} 	\\ 
&\le \sqrt{\sum\limits_{i_{1}, ..., i_{K}} (\mathcal{W}(i_{1}, ..., i_{K}))^{2}} 	\\
&= ||\nabla \mathcal{L}(\mathcal{W}) ||_{F}. 
\end{split}
\end{equation}
\end{proof}

Besides, the LR-TT-DNN model can be stacked on top of convolutional layers to construct a hybrid model, namely CNN+(LR-TT-DNN). Therefore, the CNN+(LR-TT-DNN) model is generated from the low-rank TT decomposition applied to the well-trained FC layers. The CNN framework is composed of a series of convolutional layers and flexibly deals with inputs with the formats of vectors, tensors, and time-series signals.

\section{Experiments}
\label{sec6}

Our experiments are conducted on speech enhancement and SCR tasks, and our goal is to assess the performance of the LR-TT-DNN and CNN+(LR-TT-DNN) models with empirical results. The experiments of speech enhancement are carried out within a regression framework with deep models. On the other hand, the SCR task is conducted to evaluate the classification performance. The goals of our experiments are designed as follows:
\begin{enumerate}
\item Examine the benefits of the small TT-ranks in the TT-DNN model. 
\item Examine the advantages of the hybrid models with the CNN framework with and without the small TT-ranks applied. 
\item Compare the CNN+(TT-DNN) model with the CNN+(LR-TT-DNN) one with the same small TT-ranks but with different initialization. 
\end{enumerate}

\subsection{Experiments of Speech Enhancement}
\subsubsection{Data profile} 

The experiments of speech enhancement were conducted on the Edinburgh noisy speech database~\cite{valentini2016speech}, where the noisy backgrounds of the training data are different from the test ones. More specifically, clean utterances were recorded from $56$ speakers including $28$ males and $28$ females from different accent regions of both Scotland and the United States. The clean utterances were randomly split into $23075$ training and $824$ test waveforms, respectively. The noisy training waveforms at $4$ SNR levels, (i.e., $15dB$, $10dB$, $5dB$, and $0dB$), were created by corrupting clean utterances with $10$ noisy types, which resulted in $40$ different noisy backgrounds for the training dataset. As for the test dataset, $5$ additional noise types were included to generate the test dataset. Hence, the noisy conditions for the training dataset were unmatched with the test one in terms of different noise types at various SNR values. 

\subsubsection{Experimental setup} 

In our experiments, we used $256$-dimensional normalized log-power spectral (LPS)~\cite{deng2010binary} features to represent the speech signals. The LPS features were generated by applying a $512$-point fast Fourier transform (FFT) on a speech segment of $32$ms. For each input frame, $17$ adjacent neighboring frames were concatenated into a high dimensional feature vector, which results in $17\times 256$ input dimensions. The clean reference features associated with the noisy inputs were assigned to the top layer during the training process, and the enhanced speech features are taken as the output in the test process. Besides, the overlap-add method was used to generate the enhanced speech signals~\cite{crochiere1980weighted}. The criterion of minimum absolute error (MAE)~\cite{chai2014root} is taken as the objective loss function during the training process and it is also used to measure the test performance on the test dataset. Our previous work~\cite{mae_spl} analyzed the performance advantage of MAE over mean squared error (MSE)~\cite{wallach1989mean} and root mean squared error (RMSE)~\cite{willmott2005advantages} for DNN-based vector-to-vector regression. Besides, the Adam optimizer~\cite{adam} with a learning rate of $0.001$ was used in the training stage. 

\begin{figure}[htp]
\centerline{\epsfig{figure=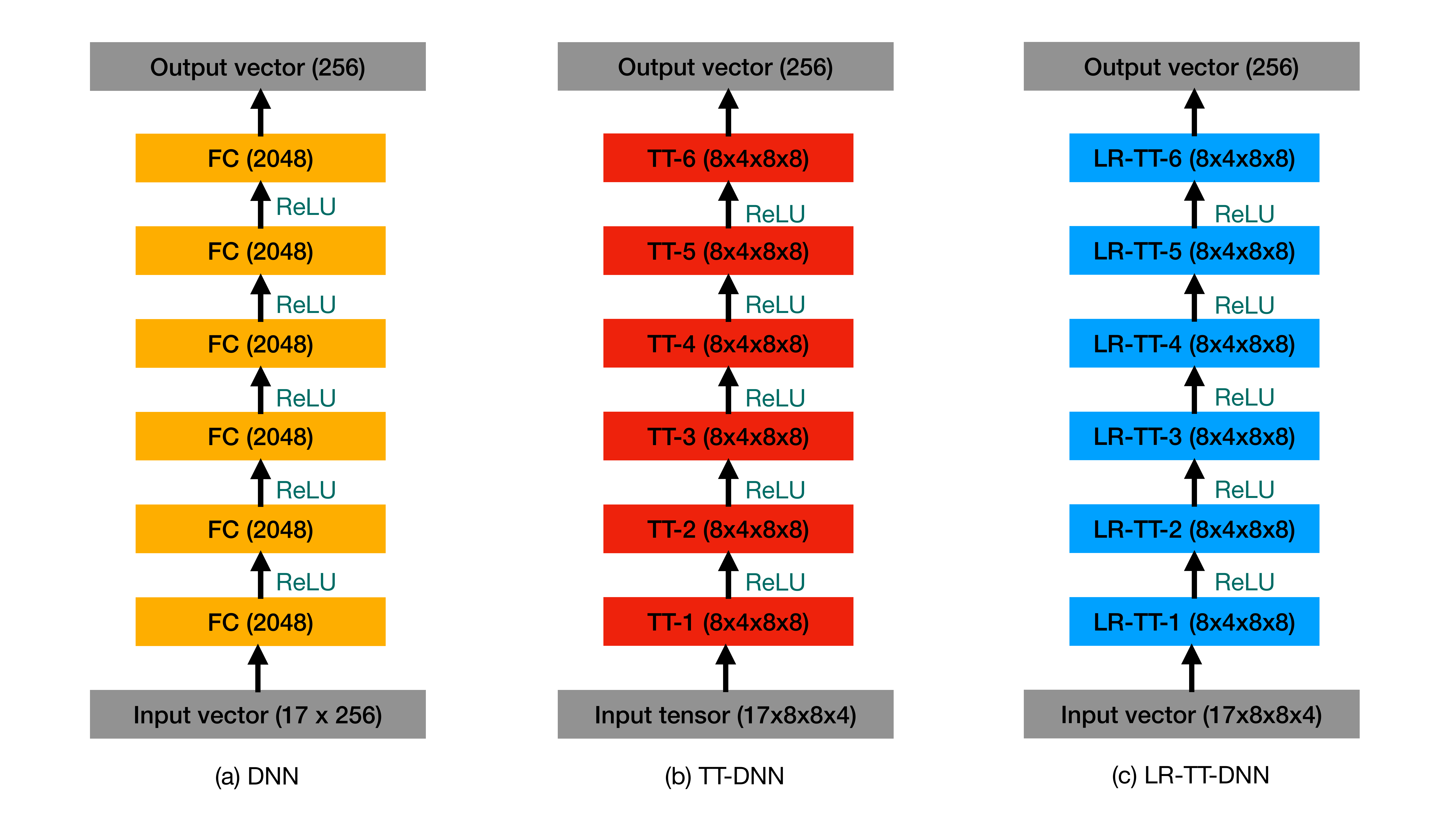, width=95mm}}
\caption{{\it The architectures of DNN, TT-DNN, and LR-TT-DNN models for speech enhancement.}}
\label{fig:exp1}
\end{figure}

The following neural architectures were first investigated: DNN, TT-DNN, and LR-TT-DNN, as shown in Figure~\ref{fig:exp1}. The DNN owned $6$ hidden layers with $2048$ units each, and both TT-DNN and LR-TT-DNN consisted of $6$ hidden TT layers. The TT-DNN model was trained from scratch and the LR-TT-DNN model was built from that. Moreover, the prescribed large and small TT-ranks were separately set as $\{1, 12, 12, 12, 1\}$ and  $\{1, 3, 4, 3, 1\}$, respectively. Concerning the TT layers in Figure~\ref{fig:exp1}, each TT layer is set as $4$-order tensor format of $8\times 4\times 8\times 8$. Accordingly, each FC hidden layer in the DNN has $2048$ units with the ReLU activation.

\begin{figure}[htp]
\centerline{\epsfig{figure=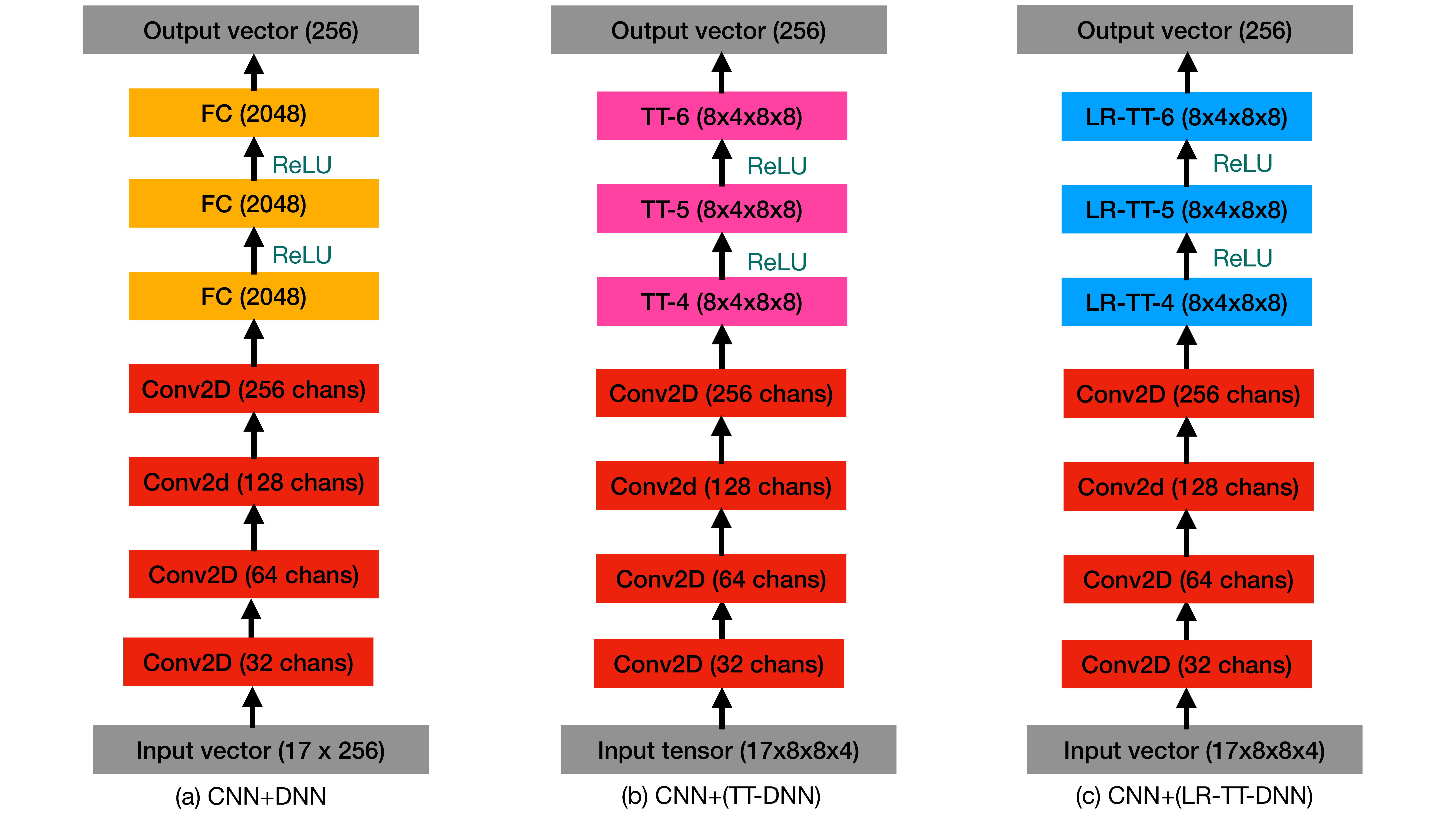, width=95mm}}
\caption{{\it The architectures of CNN+DNN, CNN+(TT-DNN), and CNN+(LR-TT-DNN) models for speech enhancement.}}
\label{fig:exp2}
\end{figure}

Then, we discuss the hybrid architectures combining convolutional neural networks with the FC or TT-DNN layers, as shown in Figure~\ref{fig:exp2}. There are four 2D convolutional layers stacked at the bottom, and three FC or TT hidden layers placed at the top of the convolutional layers outputting the regression outcomes. 


Perceptual Evaluation of Speech Quality (PESQ)~\cite{rix2001perceptual} and short-time objective intelligibility (STOI)~\cite{taal2010short} metrics were adopted to evaluate the speech quality. PESQ, which lies in the domain from $-0.5$ to $4.5$, is an indirect assessment and a higher PESQ score means a better-perceived quality of speech signals. Similarly,  STOI  ranges from $0$ to $1$ and refers to a measurement of predicting speech intelligibility. The higher the value of the STOI score is, the better the intelligibility of the speech signal is.

\subsubsection{Experimental results}

Table~\ref{tab:tab1} presents the results of the DNN, TT-DNN, and LR-TT-DNN models in Figure~\ref{fig:exp1}. The baseline based on the DNN structure can attain an outstanding speech enhancement performance, but the number of model parameters is as high as 30Mb. In comparison, the TT-DNN model owns much fewer model parameters than the DNN one (0.604Mb vs. 30.425Mb) and achieves a lower MAE score (0.664 vs. 0.675). However, the TT-DNN model performs worse than the DNN one in terms of PESQ (2.76 vs. 2.78) and STOI (0.853 vs. 0.856) scores. On the other hand, the LR-TT-DNN model owns fewer model parameters than the DNN (0.531 vs. 30.425) and the TT-DNN (0.531 vs. 0.604) models and performs consistently better than the DNN baseline and the TT-DNN models.

\begin{table}[tpbh]\footnotesize
\center
\renewcommand{\arraystretch}{1.3}
\caption{The experimental results on the test dataset for speech enhancement using DNN, TT-DNN, and LR-TT-DNN models.}
\begin{tabular}{|c||c|c|c|c|}
\hline
Models     		& Params	$(\text{Mb})$	         &   MAE     	   &    PESQ        &   STOI   		\\
\hline
DNN	        		&  30.425					&   0.675	   	   &	2.78		  &    0.856		\\
\hline
TT-DNN    	&  0.604					&   0.664   	   &	2.76	 	  &    0.853		\\
\hline
LR-TT-DNN	& 0.531 	    				&   0.601	       	   &	2.78	 	  &    0.859		\\
\hline
\end{tabular}
\label{tab:tab1}
\end{table}

Table~\ref{tab:tab2} shows the empirical results of the models in Figure~\ref{fig:exp2}. The hybrid model CNN+DNN can significantly improve the DNN baseline with a lower MAE score (0.614 vs. 0.675), a much higher PESQ score (3.13 vs. 2.78), and a higher STOI score (0.887 vs. 0.856). Although both CNN+(TT-DNN) and CNN+(LR-TT-DNN) own more model parameters than the TT-DNN and LR-TT-DNN counterparts, the CNN+(LR-TT-DNN) model obtains the lowest MAE score and the highest PESQ and STOI scores. All these results highlight the advantage of using small TT-ranks for TT-DNN in speech enhancement. 

\begin{table}[tpbh]\footnotesize
\center
\renewcommand{\arraystretch}{1.3}
\caption{The experimental results on the test dataset for speech enhancememnt using CNN+DNN, CNN+(TT-DNN), and CNN+(LR-TT-DNN) models.}
\begin{tabular}{|c||c|c|c|c|}
\hline
Models     	  & Params	$(\text{Mb})$	   &   MAE     	  &    PESQ                &   STOI   	\\
\hline
CNN+DNN	   	  &  17.486	&   0.614  	   	   &	3.13		  &  0.887		\\
\hline
CNN+(TT-DNN)	  &   1.241		&   0.598		   &   3.11 		  &  0.883		\\
\hline
CNN+(LR-TT-DNN)   &   1.216		&   0.587 	  	   &	3.14		  &  0.889  	\\
\hline
\end{tabular}
\label{tab:tab2}
\end{table}

\begin{figure}[htp]
\centerline{\epsfig{figure=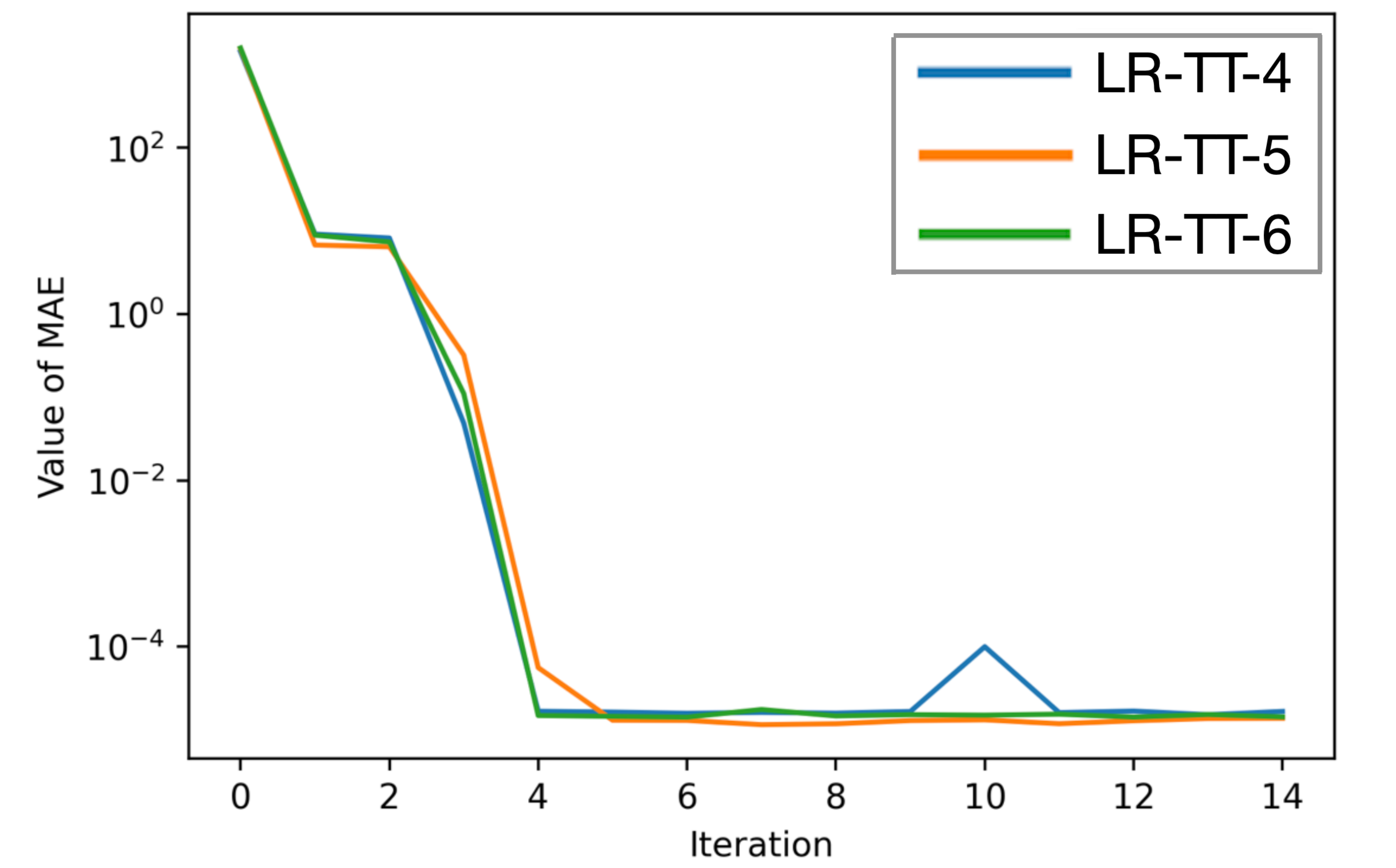, width=65mm}}
\caption{{\it Top 3 LR-TT layers of the LR-TT-DNN model.}}
\label{fig:conv1}
\end{figure}

\begin{figure}[htp]
\centerline{\epsfig{figure=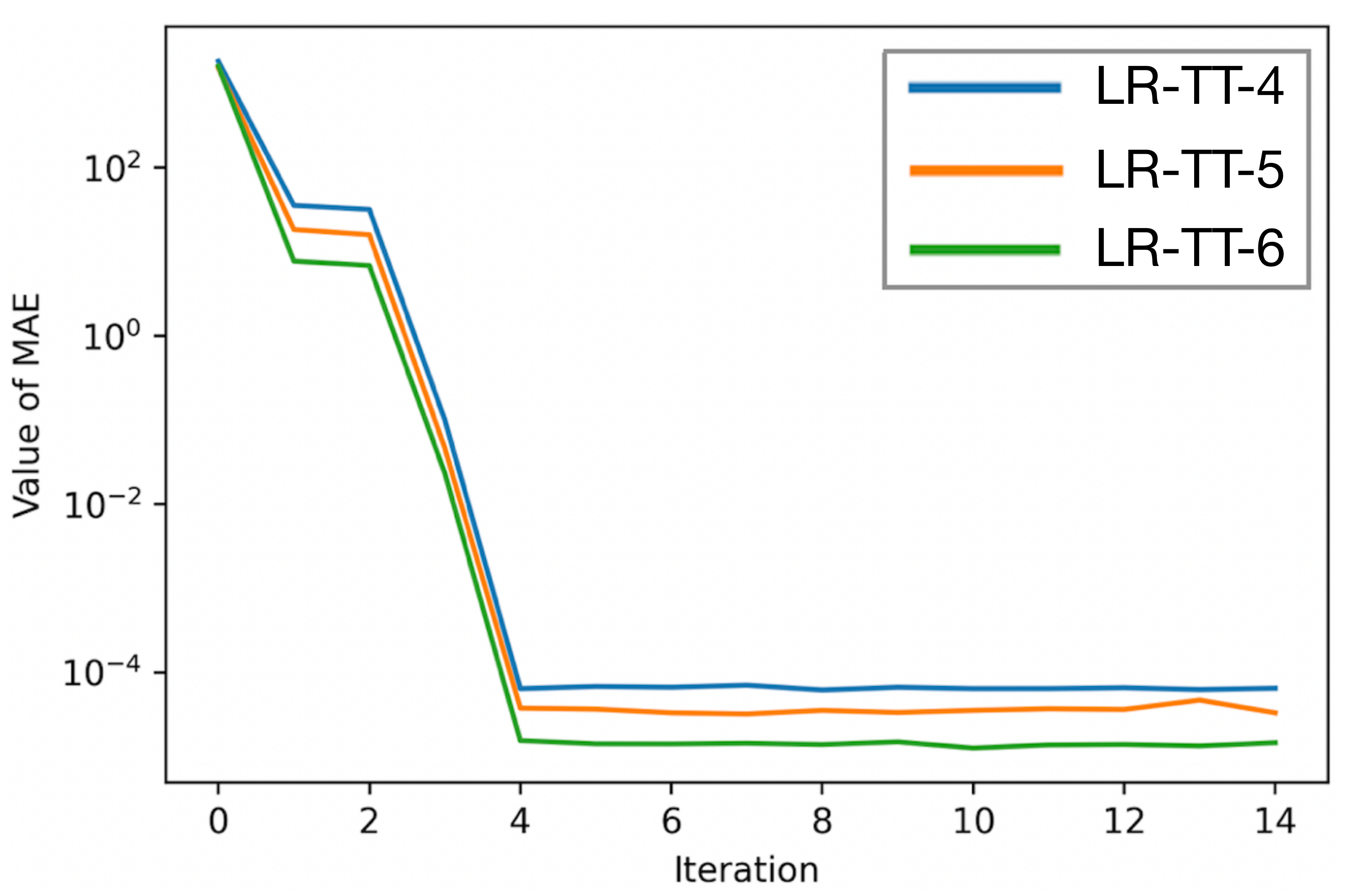, width=65mm}}
\caption{{\it Top 3 LR-TT layers of the CNN+(LR-TT-DNN) model.}}
\label{fig:conv2}
\end{figure}

Besides, Figure~\ref{fig:conv1} and~\ref{fig:conv2} separately illustrate the convergence curves of the RGD algorithm for low-rank TT decomposition to the TT layers of the TT-DNN model. The curves demonstrate that the LR-TT layers get convergence after $4$ iterations and the LR-TT layers at the top may exhibit lower approximation error compared with the ones at the bottom. 

Finally, by employing the same small TT-ranks $\{1, 3, 4, 3, 1\}$, we compare a randomly initialized CNN+(TT-DNN) model with the CNN+(LR-TT-DNN) model derived from a well-trained CNN+(TT-DNN) with large TT-ranks. The architectures of the two models are shown in Figure~\ref{fig:exp2} (b) and Figure~\ref{fig:exp2} (c), respectively, and both need further fine-tuning based on the SGD algorithm. Table~\ref{tab:rand_speech} shows the results of these two models and highlights the performance advantage of the CNN+(LR-TT-DNN) model. In more detail, the CNN+(LR-TT-DNN) model owns the same number of parameters as the CNN+(TT-DNN) one and performs better in terms of a lower MAE score ($0.612$ vs. $0.587$), higher PESQ score (3.14 vs. 3.08), and higher STOI score (0.889 vs. 0.881) than the CNN+(TT-DNN) model.

\begin{table}[tpbh]\footnotesize
\center
\renewcommand{\arraystretch}{1.3}
\caption{The experimental results on the test dataset for speech enhancememnt with the CNN+(TT-DNN) model and the CNN+(LR-TT-DNN) model with the same small TT-ranks.}
\begin{tabular}{|c||c|c|c|c|}
\hline
Models   & Params	 $(\text{Mb})$	&   MAE    &    PESQ        &   STOI   	\\
\hline
CNN+(TT-DNN)	  &   1.216		&   0.612	 &   3.08 	 	&  0.881	\\
\hline
CNN+(LR-TT-DNN)    &   1.216		&   0.587 	 &   3.14		&  0.889  	\\
\hline
\end{tabular}
\label{tab:rand_speech}
\end{table}

\subsection{Experiments of Spoken Command Recognition}
\subsubsection{Data profile}
We conduct our SCR experiments on the Google Speech Command Dataset~\cite{warden2018speech}. The dataset consists of $35$ spoken commands (e.g.,  [`left', `go', `yes', `down', `up', `on', `right', etc.), which results in $11,165$ development and $6,500$ test utterances in total. The development dataset is randomly split into two parts: $90\%$ of data are used for the model training and $10\%$ of data are taken for validation. All the audio files are downsampled from $16KHz$ to $8KHz$ and they are about $1$ second long. A mini-batch size of $256$ is set in the training process, and the speech signals in a batch are configured as the same length by using zero padding. 

\subsubsection{Experimental setup}
\begin{figure}[htp]
\centerline{\epsfig{figure=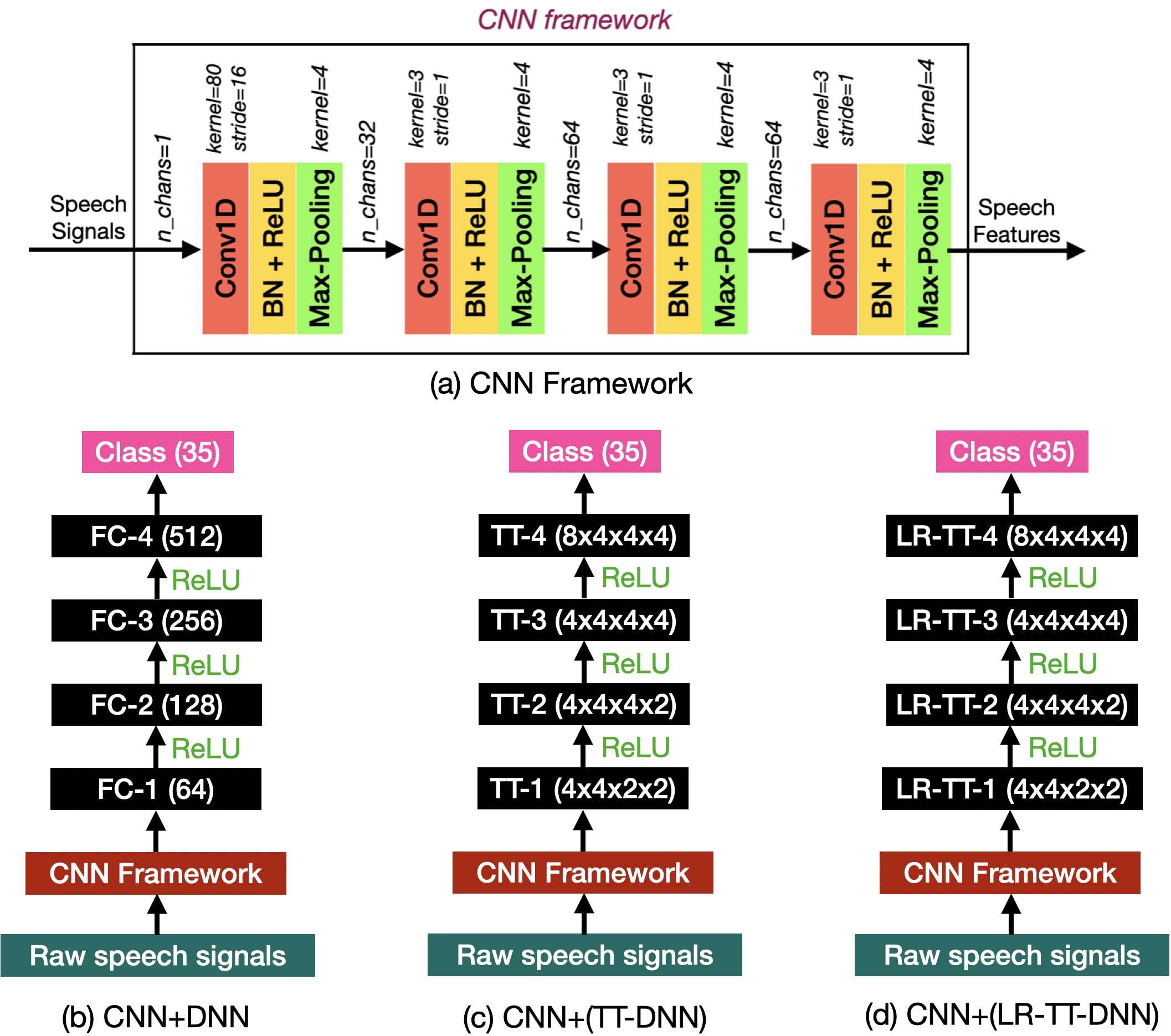, width=75mm}}
\caption{{\it CNN+DNN, CNN+(TT-DNN) and CNN+(LR-TT-DNN) models for spoken command recognition, where the tensor shape $4\times 4 \times 2 \times 2$ refers to the $4$-order tensor in the space $\mathbb{R}^{4\times 4\times 2\times 2}$.}}
\label{fig:asr}
\end{figure}

The model structures for the evaluated SCR systems are shown in Figure~\ref{fig:asr}, where three acoustic models are taken into account. Figure~\ref{fig:asr} (a) presents the CNN framework transforming the input speech signals into spectral features. The CNN framework consists of $4$ Conv1D models, each of which follows the batch normalization (BN)~\cite{santurkar2018does} and ReLU operations followed by a Max-Pooling step~\cite{murray2014generalized}. Figure~\ref{fig:asr} (b) shows the CNN+(TT-DNN) model in which $4$ FC layers ($64$-$128$-$256$-$512$) are stacked and $35$ classes are appended as the label layer. Figure~\ref{fig:asr} (c) illustrates our CNN+(TT-DNN) model in which $4$ TT layers follow the tensor shape of $4\times 4 \times 2 \times 2$ - $4\times 4 \times 4 \times 2$ - $4\times 4 \times 4 \times 4$ - $8\times 4 \times 4 \times 4$, and Figure~\ref{fig:asr} (d) shows the shape of the CNN+(LR-TT-DNN) model which is derived from the low-rank TT decomposition on the CNN+(TT-DNN) model. 

The loss objective function was set up based on the criterion of cross-entropy (CE)~\cite{zhang2018generalized}. The Adam optimizer is used for the optimizer with a learning rate of $0.01$ in the training stage. The CE and accuracy are employed to assess the model performance. CE is a direct measurement corresponding to the loss function, and accuracy is an indirect measurement to evaluate the performance of spoken command recognition. There are a total of $100$ epochs used in the training process, and the final accuracy is calculated for the SCR performance on 10 runs. 

Our experiments consist of three parts: (1) we construct the CNN+(TT-DNN) model with randomly initialized model parameters and prescribed large TT-ranks $\{1, 12, 12, 12, 1\}$; (2) we transform each well-trained TT hidden layer of the CNN+(TT-DNN) model to the resulting LR-TT layer with small TT-ranks $\{1, 3, 4, 3, 1\}$ for the CNN+(LR-TT-DNN) model; (3) we utilize Tucker decomposition on a well-trained DNN model and compare its performance with the hybrid tensor models.

\subsubsection{Experimental Results}

Our experimental results are shown in Table~\ref{tab:res3}. The CNN+DNN model is taken as the baseline of the SCR system, and it attains $94.42\%$ accuracy and 0.251 CE score, which are better than the results of the DenseNet model, the neural attention model, and the QCNN model. The CNN+(TT-DNN) model obtains better results in terms of accuracy ($96.31\%$ vs. $94.42\%$), CE ($0.137$ vs. $0.251$), and it owns a much smaller model size ($0.056$ vs. $0.216$) than the CNN+DNN model. More importantly, the CNN+(TT-DNN) model also outperforms the DenseNet, neural attention, and QCNN models in terms of smaller model size, smaller CE value, and higher classification accuracy. Furthermore, the CNN+(LR-TT-DNN) model, which owns fewer model parameters than the CNN+(TT-DNN) model ($0.043$ vs. $0.056$), obtains higher accuracy ($96.64\%$ vs. $96.31\%$) and lower CE value ($0.124$ vs. $0.137$) than the CNN+(TT-DNN) model. All these results highlight the advantage of applying small TT-ranks for TT-DNN in SCR. 

\begin{table}[tpbh]\footnotesize
\center
\caption{The experimental results on the test dataset for spoken command recognition. Params. represents the number of model parameters; CE means the cross-entropy; and Acc. refers to the classification accuracy.}
\renewcommand{\arraystretch}{1.3}
\begin{tabular}{|c||c|c|c|c|}
\hline
Models      		& Params	$(\text{Mb})$		&   CE     		& Acc.  ($\%$)   \\
\hline
DenseNet-121~\cite{mcmahan2018listening}		&   7.978		&   0.473 	 	&    82.11			\\
\hline
Attention-RNN~\cite{de2018neural}				&   0.170		&   0.291 	 	&    93.90 			\\
\hline
QCNN~\cite{yang2021decentralizing}			&   0.186		&   0.280 	 	&    94.23			\\
\hline
CNN+DNN			&   0.216		&   0.251 	 		&    94.42 			\\
\hline
CNN+(TT-DNN)		&   0.056 		&   0.137			&   96.31			\\
\hline
CNN+(LR-TT-DNN)	& \textbf{0.043}   	&  \textbf{0.124}	&  \textbf{96.64}	\\
\hline
\end{tabular}
\label{tab:res3}
\end{table}

Moreover, we compare our hybrid TT models with Tucker decomposition on the DNN. Tucker decomposition is a high-order extension to the singular value decomposition by computing the orthonormal spaces associated with the different modes of a tensor. Moreover, the CP decomposition can be taken as a special case of Tucker decomposition. Thus, it is meaningful to verify whether Tucker decomposition applied to each FC layer of the DNN can lead to a model parameter reduction with only a small drop in performance.  

The Tucker decomposition to the well-trained CNN+DNN baseline model is denoted as CNN+(TD-DNN) model. Table~\ref{tab:tab5} compares the results of the CNN+(TD-DNN) model with those of the CNN+DNN, CNN+(TT-DNN), and CNN+(LR-TT-DNN) models. Although the CNN+(TD-DNN) model can slightly decrease the number of model parameters of the CNN+DNN, the SCR performance can be significantly degraded in terms of higher CE (0.286 vs. 0.251) and much lower accuracy ($91.25\%$ vs. $94.42\%$). Moreover, the results of the CNN+(TD-DNN) model are consistently worse than those of the CNN+(TT-DNN) and CNN+(LR-TT-DNN) models. Our results suggest that Tucker decomposition on DNN cannot maintain the baseline performance of the TT technique. 

\begin{table}[tpbh]\footnotesize
\center
\renewcommand{\arraystretch}{1.3}
\caption{A comparison of the CNN+(TD-DNN) model with the CNN+(LR-TT-DNN), CNN+(TT-DNN) and CNN+DNN models on the test dataset for spoken command recognition. }
\begin{tabular}{|c||c|c|c|c|}
\hline
Models      			& Params	$(\text{Mb})$	&   CE     	&    Acc.  ($\%$)     			 \\
\hline
CNN+DNN			&   0.216		&   0.251 	 	&   94.42 			\\		
\hline
CNN+(TD-DNN)		&   0.206		&   0.286		&   91.25			\\
\hline
CNN+(TT-DNN)		&   0.056		&   0.137		&   96.31			\\
\hline
CNN+(LR-TT-DNN)		&   \textbf{0.043}           &   \textbf{0.124}		&   \textbf{96.64}			\\	
\hline
\end{tabular}
\label{tab:tab5}
\end{table}

Finally, by using the same small TT-ranks $\{1, 3, 4, 3, 1\}$, we compare a randomly initialized CNN+(TT-DNN) model with the CNN+(LR-TT-DNN) model derived from a well-trained CNN+(TT-DNN) with large TT-ranks. The two models follow the architectures as shown in Figure~\ref{fig:asr} (c) and Figure~\ref{fig:asr} (d), respectively, and both need further fine-tuning based on the SGD algorithm. Table~\ref{tab:rand_speech2} presents the related empirical results, where it is seen that the CNN+(LR-TT-DNN) model outperforms the CNN+(TT-DNN) one in terms of lower CE score and higher accuracy.

\begin{table}[tpbh]\footnotesize
\center
\renewcommand{\arraystretch}{1.3}
\caption{A comparison of the CNN+(TT-DNN) model and the CNN+(LR-TT-DNN) model with the same TT-ranks for the test dataset for spoken command recognition.}
\begin{tabular}{|c||c|c|c|}
\hline
Models   			  & Params $(\text{Mb})$	&   CE	   	&  Acc. ($\%$)      	\\
\hline
CNN+(TT-DNN)	  &   0.043				&   0.135	 	&   96.17 	 		\\
\hline
CNN+(LR-TT-DNN)    &    0.043			&   0.124	  	&   96.64			\\
\hline
\end{tabular}
\label{tab:rand_speech2}
\end{table}

\section{Conclusions}
\label{sec7}
This work has presented two novel TT models, namely LR+(TT-DNN) and CNN+(LR-TT-DNN), whose performance has been measured in speech enhancement and SCR tasks. We first introduce the RGD algorithm to generate an LR-TT-DNN model from a well-trained TT-DNN model, and then we analyze the theoretical performance of the LR-TT-DNN model. Then, we construct the hybrid model with a CNN to generate the CNN+(LR-TT-DNN) model. Our experiments on speech enhancement and SCR tasks suggest that our proposed LR-TT-DNN model with a smaller size achieves better performance than the DNN and TT-DNN counterparts and that the hybrid model combining LR-TT-DNN with CNN achieves the best performance.

\appendices
\section{Definition of a $\tau$-gradient dominated function}
\label{app1}
\begin{definition}
For a closed convex set $\mathcal{C} \subset \mathbb{R}^{D}$ and function $f$ that is differentiable on an open set containing $\mathcal{C}$, we say $f$ is $\tau$-gradient dominated over $\mathcal{C}$ if there exists a constant $\tau \ge 0$ such that $\forall \textbf{x}, \textbf{x}' \in \mathcal{C}$, we have
\begin{equation*}
\min\limits_{\textbf{x}' \in \mathcal{C}} f(\textbf{x}') \ge f(\textbf{x}) + \min\limits_{\textbf{x}' \in \mathcal{C}} \left[ \langle \nabla f(\textbf{x}), \textbf{x}' - \textbf{x} \rangle + \frac{\tau}{2}\left|\left| \textbf{x} - \textbf{x}' \right|\right|_{2}^{2} \right]. 
\end{equation*}
\end{definition}

\ifCLASSOPTIONcaptionsoff
  \newpage
\fi

\bibliographystyle{IEEEtran}
\bibliography{speech}

\begin{IEEEbiography}[{\includegraphics[width=1.1in,height=1.35in,clip,keepaspectratio]{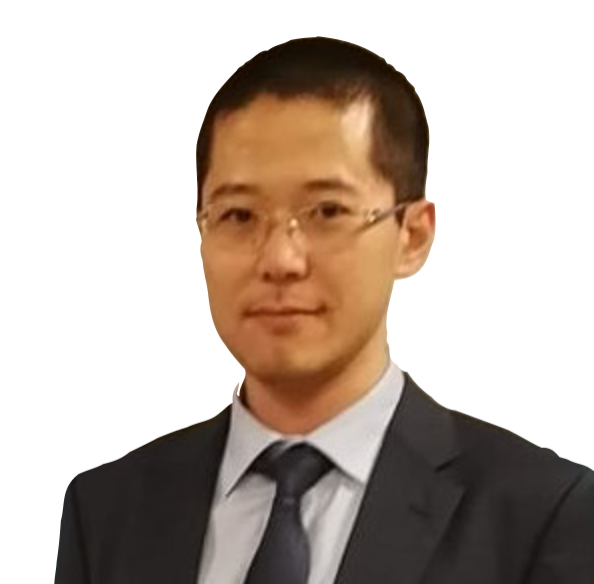}}]{Jun Qi}
received the Ph.D. degree in the School of Electrical and Computer Engineering from Georgia Institute of Technology, Atlanta, USA, in 2022. Besides, he received a Master's degree in Electrical Engineering from the University of Washington, Seattle, USA, in 2017, and a Master's degree in Electronic Engineering from Tsinghua University, Beijing, China, in 2013. Previously, he was a researcher at Microsoft Research, Redmond, USA, in 2017. His research concentrates on (1) Quantum Tensor Network in Machine Learning and Signal Processing; (2) Quantum Neural Networks for Speech and Language Processing; (3) Quantum Cryptography and Quantum Optimization Algorithms. He was the recipient of 1st prize in Xanadu AI Quantum Machine Learning Competition, 2019, and his ICASSP paper about quantum speech recognition was nominated as the best paper candidate in 2022. Besides, he gave tutorials on Quantum Neural Networks for Speech and Language Processing at IJCAI'21 and another tutorial at ICASSP'22. 
\end{IEEEbiography}

\begin{IEEEbiography}[{\includegraphics[width=1.1in,height=1.35in,clip,keepaspectratio]{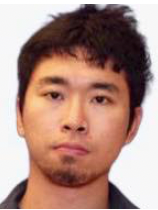}}]{Chao-Han Huck Yang}
received the B.S. degree from National Taiwan University, Taipei City, Taiwan, in 2016. He is currently pursuing a Ph.D. degree from the School of Electrical and Computer Engineering, Georgia Institute of Technology, Atlanta, GA, USA. His recent research interests include the adversarial robustness of deep neural networks and reinforcement learning with real-world applications on speech processing, communication networks, and audio-visual processing. He is a Student Member of IEEE Society. He was a recipient of the Wallace H. Coulter Fellowship from the Georgia Institute of Technology, in 2017 and 2018. He received IEEE SPS Travel Grant for ICIP 2019, 1st Prize on the Research Track of Xanadu Quantum Software Global Competition, and DeepMind Travel Award for NeurIPS 2019. He took research interns in the Image and Visual Representation Laboratory (IVRL), the Ecole Polytechnique Federale de Lausanne (EPFL), Switzerland, in 2018, KAUST, and the Amazon Alexa Research.
\end{IEEEbiography}

\begin{IEEEbiography}[{\includegraphics[width=1.1in,height=1.35in,clip,keepaspectratio]{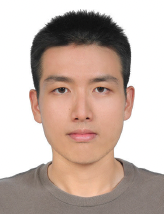}}]{Pin-Yu Chen}
received the B.S. degree (Hons.) in electrical engineering and computer science from National Chiao Tung University, Taiwan, in 2009, the M.S. degree in communication engineering from National Taiwan University, Taiwan, in 2011, and the M.A. degree in Statistics and the Ph.D. degree in electrical engineering and computer science from the University of Michigan, Ann Arbor, USA, in 2016. He is currently a Research Staff Member with the IBM Thomas J. Watson Research Center, Yorktown Heights, NY, USA. He is also the Chief Scientist of RPI-IBM AI Research Collaboration and a PI of ongoing MIT-IBM Watson AI Lab projects. His recent research is on adversarial machine learning and the robustness of neural networks. His long-term research vision is building trustworthy machine learning systems. He has published more than 20 articles on trustworthy machine learning at major AI and machine learning conferences, given tutorials at CVPR'20, ECCV'20, ICASSP'20, KDD'19, and Big Data'18, and co-organized several workshops on adversarial learning for machine learning and data mining. His research interests also include graph and network data analytics and their applications to data mining, machine learning, signal processing, and cyber security. At IBM Research, he has co-invented more than 20 U.S. patents. In 2019, he received two Outstanding Research Accomplishments on research in adversarial robustness and trusted AI, and one Research Accomplishment on research in graph learning and analysis. He was a recipient of the Chia-Lun Lo Fellowship from the University of Michigan, Ann Arbor. He received the NIPS 2017 Best Reviewer Award and was also a recipient of the IEEE GLOBECOM 2010 GOLD Best Paper Award. He is also on the editorial board of PLOS One.
\end{IEEEbiography}

\begin{IEEEbiography}[{\includegraphics[width=1.1in,height=1.35in,clip,keepaspectratio]{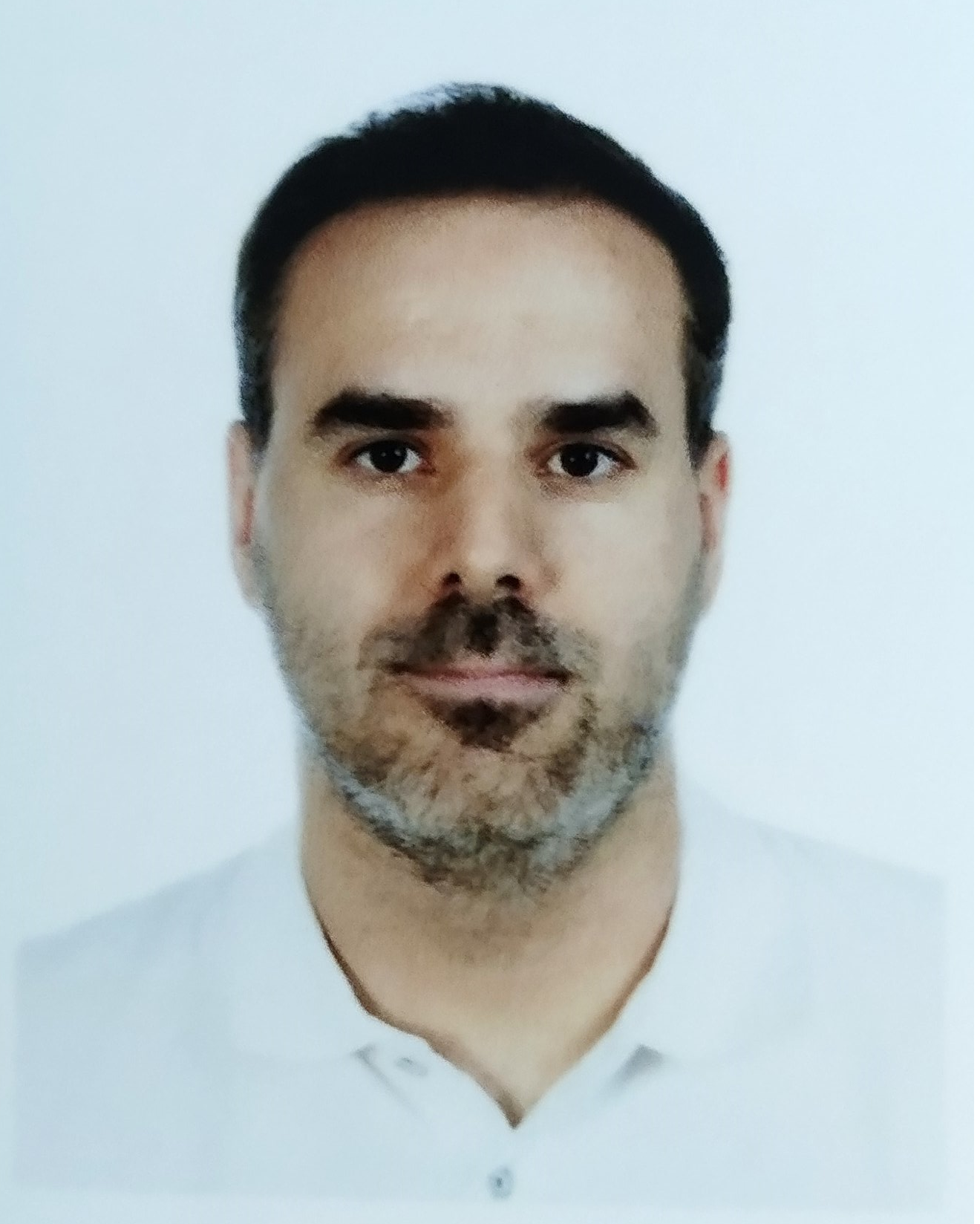}}]{Javier Tejedor}
Javier Tejedor received the B.Sc. degree in computer engineering and the M.Sc. and Ph.D. degrees in computer and telecommunication engineering from Universidad Autónoma de Madrid, Madrid, Spain, in 2002, 2005, and 2009 respectively. From 2001 to 2013, he was with Human-Computer Technology Laboratory (HCTLab), Universidad Autónoma de Madrid. From 2003 to 2013, he was an Assistant Professor with the School of Computer Engineering and Telecommunication. From 2014 to 2016, he was with GEINTRA, Universidad de Alcalá, Madrid, Spain, as an Associate Researcher. Since 2016, he has been an Assistant Professor at Universidad San Pablo CEU, Madrid, Spain. His main interests are speech indexing and retrieval, spoken term detection, large vocabulary continuous speech recognition, and pattern recognition applied to pipeline monitoring systems and biomedical signals. 
\end{IEEEbiography}







\end{document}